\documentclass[journal]{IEEEtran}

\makeatletter
\def\ps@headings{%
	\def\@oddhead{\mbox{}\scriptsize\rightmark \hfil \thepage}%
	\def\@evenhead{\scriptsize\thepage \hfil \leftmark\mbox{}}%
	\def\@oddfoot{}%
	\def\@evenfoot{}}
\makeatother 

\makeatletter
\def\endthebibliography{%
	\def\@noitemerr{\@latex@warning{Empty `thebibliography' environment}}%
	\endlist
}
\makeatother

\usepackage{graphicx}
\usepackage{epstopdf}
\usepackage{amsmath, amsfonts, epsfig, multirow, floatflt}
\usepackage{amssymb}
\usepackage{subfigure}
\usepackage{cite}
\usepackage{algorithm}
\usepackage{algorithmicx}
\usepackage{algpseudocode}
\usepackage{hyperref}
\usepackage{enumitem}
\usepackage{diagbox}
\usepackage{mathrsfs}
\usepackage{url}
\usepackage{color}

\usepackage{caption}
\usepackage{subcaption}
\usepackage{hhline}
\usepackage{array}
\usepackage{booktabs}
\usepackage{amsthm}
\usepackage{bm}
\usepackage[table,xcdraw]{xcolor}
\usepackage{verbatim}

\algblock{Input}{EndInput}
\algnotext{EndInput}
\algblock{Output}{EndOutput}
\algnotext{EndOutput}

\newcommand{\Descb}[2]{\State \makebox[6em][l]{#1}#2}
\newcommand{\Descc}[2]{\State \makebox[10em][l]{#1}#2}
\newcommand{\Descd}[2]{\State \makebox[13em][l]{#1}#2}

\allowdisplaybreaks[4]
\graphicspath{{figure/}}

\begin{document}
	
 \title{Optimizing Electric Bus Charging Scheduling with Uncertainties Using Hierarchical Deep Reinforcement Learning}
	
	\author{Jiaju Qi$^{1}$, Lei~Lei$^{1}$, {\it Senior Member, IEEE}, Thorsteinn Jonsson$^{2}$, Dusit Niyato$^{3}$, {\it Fellow, IEEE }
    \thanks{$^{1}$J. Qi and L. Lei are with the School of Engineering, University of Guelph, Guelph, ON N1G 2W1, Canada, {\tt\small leil@uoguelph.ca} }
    \thanks{$^{2}$T. Jonsson is with EthicalAI, Waterloo, ON N2L 0C7, Canada. }
    \thanks{$^{3}$D. Niyato is with the College of Computing and Data Science, Nanyang Technological University, 50 Nanyang Avenue, Singapore 639798. }
    }
	
	\maketitle
	
	\begin{abstract}
The growing adoption of Electric Buses (EBs) represents a significant step toward sustainable development. By utilizing Internet of Things (IoT) systems, charging stations can autonomously determine charging schedules based on real-time data. However, optimizing EB charging schedules remains a critical challenge due to uncertainties in travel time, energy consumption, and fluctuating electricity prices. Moreover, to address real-world complexities, charging policies must make decisions efficiently across multiple time scales and remain scalable for large EB fleets. In this paper, we propose a Hierarchical Deep Reinforcement Learning (HDRL) approach that reformulates the original Markov Decision Process (MDP) into two augmented MDPs. To solve these MDPs and enable multi-timescale decision-making, we introduce a novel HDRL algorithm, namely Double Actor-Critic Multi-Agent Proximal Policy Optimization Enhancement (DAC-MAPPO-E). Scalability challenges of the Double Actor-Critic (DAC) algorithm for large-scale EB fleets are addressed through enhancements at both decision levels. At the high level, we redesign the decentralized actor network and integrate an attention mechanism to extract relevant global state information for each EB, decreasing the size of neural networks. At the low level, the Multi-Agent Proximal Policy Optimization (MAPPO) algorithm is incorporated into the DAC framework, enabling decentralized and coordinated charging power decisions, reducing computational complexity and enhancing convergence speed. Extensive experiments with real-world data demonstrate the superior performance and scalability of DAC-MAPPO-E in optimizing EB fleet charging schedules.

	\end{abstract}

	
	\section{Introduction}

In recent years, the global shift toward sustainable transportation has highlighted the importance of adopting Electric Buses (EBs) as an approach for reducing urban pollution, curbing greenhouse gas emissions, and enhancing the comfort of public transit systems \cite{8910356, ZHANG2025104584}. As the deployment of EBs continues to grow, minimizing charging costs has become a critical concern for transit operators. At the same time, to better manage electricity demand, power utilities have introduced dynamic pricing models that feature real-time electricity tariffs \cite{7154489}. Leveraging Internet of Things (IoT) systems, these advancements enable bus companies to design efficient charging schedules that minimize costs. This is achieved by strategically aligning charging activities with periods of low electricity price and, when possible, supplying energy back to the grid in Vehicle-to-Grid (V2G) mode during high-price periods \cite{chandra2025optimal}. This approach gives rise to new challenges for the EB Charging Scheduling Problem (EBCSP).

In general, the EBCSP involves managing one or more EBs, a set of scheduled trips, and the associated charging infrastructure. The objective is to optimize the charging schedule to minimize charging and operational costs while ensuring that the EBs have sufficient battery energy to complete their assigned trips. This must be achieved while satisfying various operational constraints, such as adherence to bus schedules and accommodating limitations in charger availability. \par

In the literature on the EBCSP, most studies have focused on system models with deterministic and constant parameter values, such as \cite{he2023battery,manzolli2022electric}. Optimal policies in these models are typically obtained by solving a Mixed Integer Linear Programming (MILP) problem. However, while deterministic models simplify both problem formulation and solution processes, they fail to account for two key types of uncertainties in real-world scenarios: (i) uncertainties in EB operations, such as random variations in travel time and energy consumption; and (ii) uncertainties in the smart grid, such as time-varying electricity prices. Therefore, the system models that incorporate these uncertainties and stochastic elements provide a more accurate reflection of reality, leading to more reliable and efficient charging schedules.

Although the uncertainty of electricity prices is rarely addressed in existing studies on the EBCSP, recent research has begun to account for uncertainties in EB travel time and energy consumption. For instance, \cite{liu2022robust} adopted a Robust Optimization (RO) approach, while \cite{liu2022optimal} and \cite{bie2021optimization} applied Immune Algorithm (IA) and Genetic Algorithm (GA), respectively. While these algorithms effectively address challenges associated with the uncertainty in EB operations, they typically operate offline and require the entire algorithm to be re-run to integrate new information or accommodate changes in the environment.  \par 

Combining Reinforcement Learning (RL) with Deep Neural Networks (DNNs), Deep Reinforcement Learning (DRL) shows considerable promise for addressing uncertainties in dynamic operational environments. Unlike the previously mentioned methods, DRL learns directly from interactions with the environment, eliminating the need for a predefined model of the variables \cite{10107374}. Moreover, DRL can dynamically learn and update policies in real time, allowing it to efficiently adapt to changes. \par 

However, there is currently a paucity of literature on solving the EBCSP using DRL. In the existing DRL-based studies \cite{WANG2024103516,yan2024mixed,9014160}, well-known algorithms such as Double Deep Q Network (DDQN), Soft Actor-Critic (SAC), and Deep Deterministic Policy Gradient (DDPG) have been applied to optimize EB charging schedules. While these studies make valuable contributions, they do not fully address the following challenges that this paper seeks to tackle:

\begin{itemize}
    \item \textbf{Learning across multiple levels of temporal abstraction}: Making sequential decisions for charging schedules involves selecting actions across different time scales. For instance, charging power decisions are ideally made at a finer time scale, such as every few minutes, to account for fluctuating electricity prices, while both charger allocation and trip assignment decisions can operate on a coarser time scale, responding to the arrival or departure of EBs at the bus terminal. In the context of formulating a DRL model that achieves effective exploration of different policies and fast convergence during the learning process, it is crucial to consider the multitimescale nature of the EBSCP.
    \item \textbf{Scalability to a large-scale fleet}: Unlike approaches such as MILP, data-driven DRL-based approaches do not face scalability constraints related to the number of trips involved in EB daily operations. However, as the number of EBs increases, the state space and action space grow exponentially, leading to greater computational complexity and challenges in algorithm convergence. This is especially true when employing a comprehensive decision-making framework that simultaneously optimizes charging power, charger allocation, and trip assignment policies. Addressing the scalability issue is essential for ensuring the practical applicability of the proposed solutions in large-scale EB fleet.


\end{itemize}

To address the aforementioned challenges and capitalize on the strengths of DRL, this paper proposes a novel Hierarchical DRL (HDRL) algorithm, termed Double Actor-Critic Multi-Agent Proximal Policy Optimization Enhanced (DAC-MAPPO-E), to effectively solve the EBCSP. The primary contributions of this work are summarized as follows: 

\begin{enumerate}
    \item \textbf{HDRL Model}: Using the hierarchical architecture of Double Actor-Critic (DAC) \cite{zhang2019dac}, the original MDP is reformulated into two augmented MDPs. The high-level MDP addresses charger allocation and trip assignment decisions, while the low-level MDP focuses on determining the adjustable charging power for each EB. The HDRL model facilitates learning policies across two levels of temporal abstraction, with high-level decisions remaining effective for variable time periods and low-level decisions made at each time step. Unlike conventional MDPs, which make all decisions on the same time scale, the HDRL model leverages different temporal abstractions to attain a simpler and more efficient understanding of the environment. 
    \item \textbf{HDRL Algorithm}: We develop an HDRL algorithm to simultaneously learn both high-level and low-level policies. To address the scalability challenges of the DAC algorithm when applied to large-scale EB fleets, we propose the following enhancements:
\begin{itemize}
\item At the low level, we incorporate the Multi-Agent PPO (MAPPO) algorithm \cite{yu2022surprising} into the DAC framework. Utilizing the Centralized Training Decentralized Execution (CTDE) framework, each EB, acting as a decentralized low-level agent, makes local decisions on its charging power in coordination with other EBs, based on high-level global actions for charger allocation and trip assignment.  
\item At the high level, due to the mutual exclusion of local actions for each EB, we employ a centralized actor as in the original DAC framework. However, we improve the high-level actor network structure to include an agent network for each EB and a pair of mapping networks. This structure reduces computational complexity when sampling high-level actions and decreases the scale of the neural networks. Additionally, we incorporate an attention mechanism to learn the key feature from the global state for each EB, which reduces the input dimensionality for each agent network.

\end{itemize}
\end{enumerate}

The rest of the paper is organized as follows. Section II critically appraises the related works. The system model is introduced in Section III, while the MDP model is formulated in Section IV. The proposed algorithm is presented in Section V. Finally, our experiments are highlighted in Section VI, and conclusions are offered in Section VII.

\section{Related works}
\subsection{Research on EB charging scheduling problem}

The studies on EBCSP primarily utilize three charging technologies: (i) conductive charging, (ii) battery swapping, and (iii) wireless charging. Research on conductive charging scheduling is further categorized into two methods: plug-in charging and pantograph charging. These two methods are typically implemented through one of two distinct strategies: (i) depot charging, where EBs are charged overnight at bus depots using normal or slow chargers, and (ii) opportunistic charging\footnote{This terminology is synonymous with the ``opportunity charging" in \cite{zhou2024charging}.}, which employs fast chargers at terminal stations or bus stops \cite{zhou2024charging}. In the following, we primarily review the existing literature related to EBCSP using plug-in and opportunistic charging, which is the focus of our study. \par

Traditionally, the EBCSP is formulated as a MILP problem, assuming the system parameters are constant or known in advance. The MILP problems are solved using various methods, including Branch \& Price (BP) \cite{ji2023optimal}, column generation algorithms \cite{ZhangLe2021Oebf}, dynamic programming \cite{bao2023optimal}, and optimization solvers such as CPLEX \cite{manzolli2022electric,zhou2022electric,rinaldi2020mixed,he2023battery,he2020optimal}. \par

In real-world scenarios, travel time and energy consumption for a trip are inherently stochastic due to random factors such as traffic conditions and delays at intersections. Consequently, their exact values are often unavailable when solving the EBCSP. To address this issue, He \emph{et al.} \cite{he2023battery} employed a K-means clustering algorithm to predict an EB's travel time and energy consumption based on the vehicle registration number, departure time of trips, etc. These predicted values are then integrated into deterministic models to develop charging schedules. While this predictive approach improves the feasibility of deterministic models, it cannot fully eliminate prediction errors, which may lead to suboptimal charging schedules. \par

In addition to parameters related to the bus transit system, some studies have also incorporated the characteristics of the electric grid into their analyses. For example, Manzolli \emph{et al.} \cite{manzolli2022electric} investigated the potential of the V2G scheme, enabling EBs to sell electricity back to the grid. The studies of \cite{liu2022optimal, bao2023optimal, whitaker2023network, zhou2022robust} focused on the influence of Time-of-Use (TOU) electricity tariffs, leveraging time-varying electricity prices to develop optimal charging policies. However, these studies generally assume electricity prices to be known and static for predefined periods of the day. With the increasing integration of renewable energy sources (RES) into the grid, the Real-Time Energy Market (RTEM) \cite{RTEM} has gained prominence. This market allows participants to buy and sell wholesale electricity throughout the day, helping to balance real-time demand with the fluctuating supply from RES \cite{ISOEng2024}. We direct interested readers to \cite{en11112974, LE2019258} for more information on how real-time electricity prices are determined and communicated to users. As a consequence, real-time electricity prices are stochastic and uncertain, which introduces significant challenges for solving the EBCSP. This critical challenge has received limited attention in existing research. \par

Due to the limitations of deterministic models in addressing uncertainties, we mainly focus on related works based on stochastic models in the following. Specifically, we discuss and compare several representative studies across seven key aspects: (1) whether uncertainties in travel time/energy consumption and electricity prices are addressed; (2) whether the V2G mode is considered; (3) whether the charging schedules are optimized based on a predetermined EB-to-trip assignment, or through joint optimization that simultaneously addresses EB-to-trip assignment and charging schedules; (4) whether the constraint of a limited number of chargers is accounted for; (5) whether the charging power is dynamically optimized or considered a fixed value; (6) whether the multi-timescale nature of various charging schedule decisions is considered; (7) the solution algorithms used. \par

Traditionally, RO has been widely utilized in stochastic models to balance the dual objectives of minimizing charging costs and enhancing system robustness. For example, Hu \emph{et al.} \cite{hu2022joint} applied RO to optimize charging time, while Tang \emph{et al.} \cite{tang2019robust} made binary charging decisions. Notably, Tang \emph{et al.} considered the constraint of limited chargers and performed joint optimization with EB-to-trip assignment. However, both studies assumed a fixed charging power. Conversely, both Zhou \emph{et al.} \cite{zhou2022robust} and Liu \emph{et al.} \cite{liu2022robust} adopted RO and treated charging power as a decision variable. Among these, Liu \emph{et al.} further incorporated the constraint of a limited number of chargers. Their approach reformulated the EBCSP into a master problem and subproblems: the master problem addressed resource allocation, while the subproblems focused on optimizing charging schemes under the uncertainty of energy consumption. This framework effectively accounted for the multi-timescale nature of different charging schedule decisions, providing a more comprehensive solution to the EBCSP. 
Despite its advantages, RO often produces conservative policies aimed at risk avoidance, prompting exploration of more adaptive and dynamic algorithms. For example, Liu \emph{et al.} \cite{liu2022optimal} employed IA to optimize charging time for a single EB. Meanwhile, Bie \emph{et al.} \cite{bie2021optimization} modeled the probability distribution of energy consumption and adopted GA to minimize charging costs and trip departure delays, incorporating trip assignment decisions in the optimization process. 

Research on DRL-based solutions for the EBCSP remains relatively limited. Among existing studies, Chen \emph{et al.} \cite{9014160} utilized Double Q-learning (DQL) to decide the charging power for an EB upon its arrival at terminal stations, with the power level remaining constant throughout each charging session. This study only focused on a single EB, excluding considerations of limited chargers or joint optimization with trip assignment. In contrast, both \cite{WANG2024103516} and \cite{yan2024mixed} addressed entire EB fleets while incorporating EB-to-trip assignment decisions. In \cite{WANG2024103516}, Wang \emph{et al.} combined Clipped DQL with SAC to solve EB dispatching and charging scheduling problems simultaneously. However, this approach simplified charging decisions to a simple binary variable. In \cite{yan2024mixed}, Yan \emph{et al.} introduced a hybrid framework integrating DRL and MILP to optimize target SoC levels and assign service trips for EBs across multiple time scales. Specifically, a DRL agent using Twin Delayed DDPG (TD-DDPG) served as a high-level coordinator, making deadhead decisions for bus routes every $30$ minutes. Based on the number of assigned deadhead trips, detailed charging plans were generated every minute by solving a MILP in a rolling horizon fashion. However, this method assumed a fixed charging power. \par
Table \ref{related works} provides a comparative analysis of our work against the existing literature on EBCSP with stochastic models, highlighting key features. Most existing studies lack consideration of advanced grid characteristics, including V2G capabilities and uncertainty in electricity prices. Additionally, while many works address joint optimization with trip assignment or constraints on the number of chargers, only two studies, i.e., \cite{liu2022robust} and \cite{yan2024mixed}, stand out as being closely related to our approach by incorporating multi-timescale decision-making. Table \ref{related works} shows that our work integrates all the listed key features, filling the corresponding gaps in EBCSP research.

\begin{table*}
\centering
\color{black}
\caption{Contrasting this paper to the literature on EBCSP with stochastic models.}
\begin{tabular}{|l|c|c|c|c|c|c|c|c|c|} 
\hline
\textbf{Features} & \cite{liu2022optimal,hu2022joint} & \cite{bie2021optimization}&\cite{tang2019robust} &\cite{zhou2022robust} &\cite{liu2022robust} &\cite{WANG2024103516}&\cite{yan2024mixed}&\cite{9014160}& Our work  \\ 
\hline
\textbf{\begin{tabular}[l]{@{}l@{}}Uncertainty in \\electricity price \end{tabular}}&      &                                             &    &    &   &      &    &    &    \checkmark   \\ \hline
\textbf{V2G mode}&      &                                             &    &    &   &      &    & &  \checkmark        \\ \hline
\textbf{\begin{tabular}[l]{@{}l@{}}Joint optimization \\with trip assignment \end{tabular}}&      & \checkmark                                           & \checkmark  &    &   & \checkmark    & \checkmark   &  &   \checkmark      \\ 
\hline
\textbf{\begin{tabular}[l]{@{}l@{}}Limited number \\of chargers \end{tabular}}&      &                                             & \checkmark  &    & \checkmark &      & \checkmark &   & \checkmark          \\ 
\hline
\textbf{\begin{tabular}[l]{@{}l@{}}Adjustable \\ charging power\end{tabular}} &      &                                             &    & \checkmark  & \checkmark &      &   &   \checkmark  &\checkmark      \\ 
\hline
\textbf{Multi-timescale nature}&      &                                             &    &    & \checkmark  &     & \checkmark  & & \checkmark         \\ 
\hline
\textbf{DRL-based algorithm}                                                                             &      &                                             &    &    &   & \checkmark    & \checkmark  & \checkmark &  \checkmark         \\ 
\hline
\end{tabular}
\label{related works}
\end{table*}

\subsection{Research on Hierarchical Reinforcement Learning}
Existing approaches on Hierarchical RL (HRL) are primarily developed based on three foundational frameworks \cite{barto2003recent}, i.e., the option framework by Sutton \emph{et al.} \cite{sutton1999between}, MAXQ by Dietterich \emph{et al.} \cite{dietterich2000hierarchical}, and the Hierarchy of Abstract Machines (HAMs) by Parr and Russell \cite{parr1997reinforcement}. Among them, the option framework is the most widely used, where an option represents a high-level action associated with a subtask. Each option is defined by three key components: (i) an initiation condition, (ii) a low-level intra-option policy for selecting actions, and (iii) a termination probability function. \par

Option-Critic (OC), presented by Bacon \emph{et al.} \cite{bacon2017option}, represents a foundational approach for the option framework. Based on the policy gradient theorem, OC enables automated option learning using an end-to-end framework. Building on this, Zhang \emph{et al.} \cite{zhang2019dac} presented the DAC architecture, which formulates an HRL hierarchy as two parallel augmented MDPs. The high-level MDP addresses learning the policy over options and their termination conditions, while the low-level MDP focuses on learning intra-option policies. \par

DAC provides a general and flexible framework that seamlessly integrates with state-of-the-art policy optimization algorithms, such as PPO \cite{chen2018adaptive}, without requiring additional algorithmic modifications. This compatibility enables enhanced flexibility and performance. In this paper, we adopt DAC as the foundational framework for the design of our proposed algorithm.

\section{System model}
The variables and parameters used in this paper are summarized in Table \ref{notations}. We divide the time in a single day into $T$ equal-length time steps, indexed by $t\in\{0, \dots, T-1 \}$, with each time step having a duration of $\Delta t$.

\begin{table*}[t]
\centering
\caption{Notation used in this paper}
\begin{tabular}[b]{p{2.5cm}<{\raggedright}p{15.0cm}<{\raggedright}}
\hline
\textbf{Notations}&\textbf{Description}\\
\hline
\specialrule{0em}{1pt}{1pt}
\multicolumn{2}{l}{\textbf{Sets}}  \\
\specialrule{0em}{1pt}{1pt}
$\mathcal{A}_t$ & The state-dependent action space at time step $t$\\
$\mathcal{C}_{t}$/$\mathcal{C}_{i,t}$ & The global/local charging power action space at time step $t$\\
$\mathcal{I}_o $ & The initiation set of states for option $o$\\
$\mathcal{K}_{t}$ & The trip assignment action space at time step $t$\\
$\mathcal{M}$, $\mathcal{M}^{\rm lay}_{t}$ &The set of EBs, the set of EBs that are currently in the layover period at time step $t$\\
$\mathcal{O}_t$ & The state-dependent option space at time step $t$\\
$P(\mathcal{M}_t^{\rm lay})$ & The permutation of EBs in the layover period \\
$\mathcal{S}^{+}$, $\mathcal{S}^{\rm T}$/$\mathcal{S}$ & The state space, the set of terminal/non-terminal states\\
$\Omega_{t}$ & The charger allocation action space at time step $t$\\
\specialrule{0em}{1pt}{1pt}
\multicolumn{2}{l}{\textbf{Parameters}}  \\
\specialrule{0em}{1pt}{1pt}
$B_{i,t}$ & The EB status at time step $t$ for EB $i$, $0$ for operating periods and $1$ for layover periods \\
$E_{i,t}$ & The SoC level of the battery for EB $i$ at time step $t$\\
$H_t$ &  The historical electricity prices in the period spanning from time step $t-h$ up to time step $t$\\
$K,k$ & The number of trips in a day, the index of a trip\\
$k_{i,t}$ & The trip assigned to EB $i$ at time step $t$\\
$\hat{k}_t$ & The earliest upcoming trip departing after time step $t$\\
$M$, $M_t$ & The number of EBs, the number of EBs in the layover period\\
$N$ & The number of chargers\\
$\rho_t$ & The electricity price at time step $t$\\
$S_t$/$S_{i,t}$ & The global/local system state at time step $t$\\
$T$, $t$ & The number of time steps in a day, the index of a time step\\
$T^{\rm d}_{k}$, $T^\mathrm{d}_{k_{i,t}}$ & The departure time step for the $k$-th trip of the day in chronological order, the departure time step for the trip $k_{i,t}$\\ 
$T^{\rm o}_i$ & The duration of an operating period for EB $i$\\
$\Delta t$ & The duration of each time step\\
$\tau_{i,t}$ & The number of remaining time steps from time step $t$ to the departure time for layover periods, the number of time steps from the last departure time to time step $t$ for operating periods\\
\specialrule{0em}{1pt}{1pt}
\multicolumn{2}{l}{\textbf{Decision Variables}}  \\
\specialrule{0em}{1pt}{1pt}
$A_t$/$A_{i,t}$ & The global/local action at time step $t$\\
$c_t$/$c_{i,t}$ & The global/local charging/discharging power at time step $t$\\
$\omega_t$/$\omega_{i,t}$ & The global/local charging allocation action at time step $t$\\
$k_t$/$k_{i,t}$ & The global/local trip assignment action at time step $t$\\
\multicolumn{2}{l}{\textbf{Functions}}  \\
\specialrule{0em}{1pt}{1pt}
$C_{i,t}^{\mathrm{ba}}$/$C_{i,t}^{\mathrm{ch}}$/$C_{i,t}^{\mathrm{sw}}$ & The battery degradation/charging/switching cost for EB $i$\\
$C_{t}^{\mathrm{end}}$ & The penalty incurred when the SoC level in any EB's battery falls below the minimum battery capacity\\
$r(S_t, A_t)$/$r_i(S_{i,t}, A_{i,t})$ & The reward function\\
$\beta_{o_{t-1}}(S_t)$ & The termination condition of option $o_{t-1}$\\
$\varGamma_{i,t}(B_{i,t},\tau_{i,t})$ & The probability that the current period is terminated at time step $t$ for EB $i$\\
$\pi_{o_t}(c_t|S_t)$ & The intra-option policy for option $o$\\
$\mu(o_t|S_t)$ & The policy over options\\
\specialrule{0.05em}{2pt}{0pt}
\end{tabular}
\label{notations}
\end{table*}

\begin{figure}[t]
\centering
\includegraphics[width=0.8\linewidth]{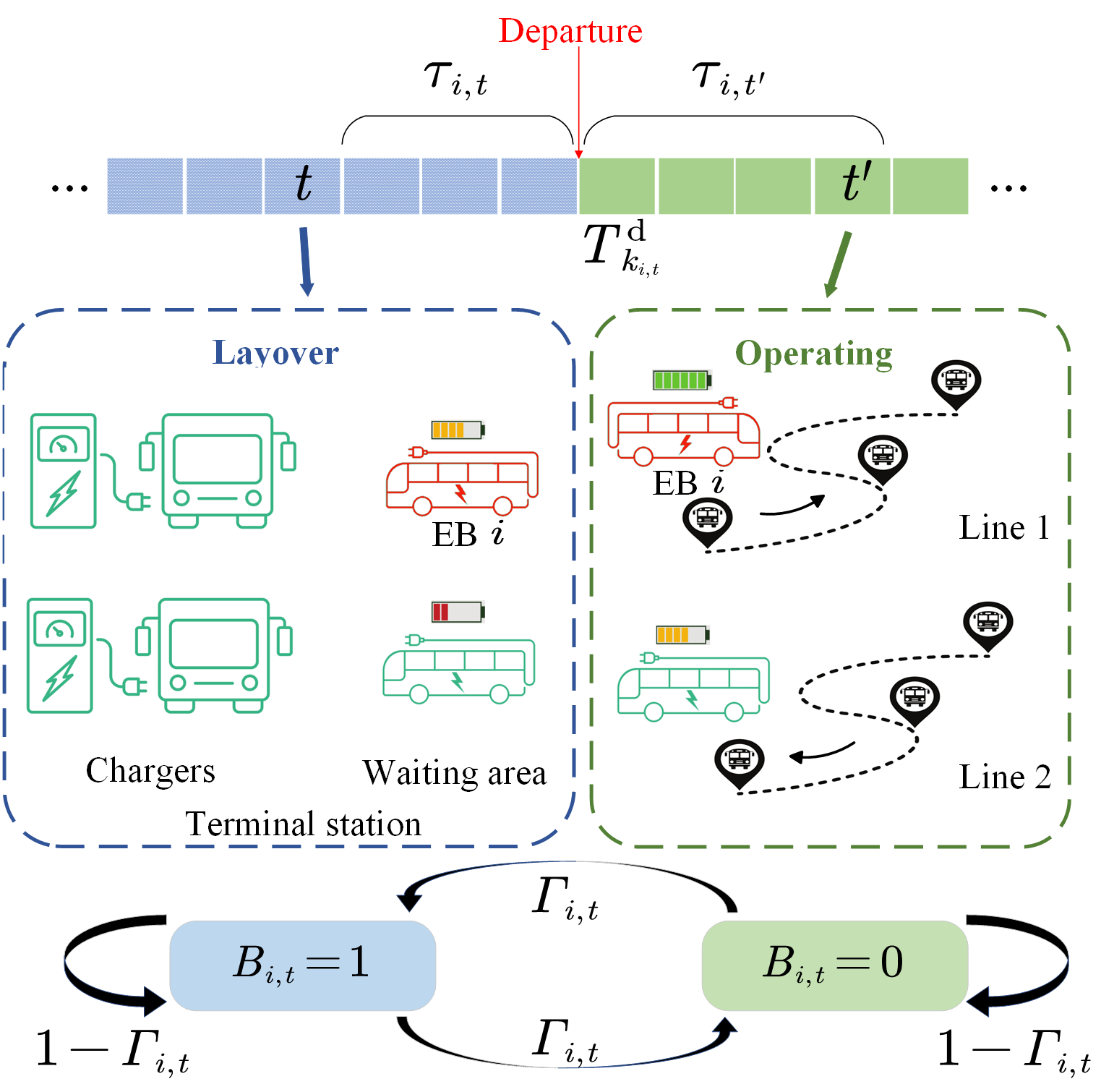}
\caption{The schematic diagram of the system model.}
\label{systemmodel} 
\end{figure}

\subsection{EB operation model}
We consider multiple bus lines sharing a single terminal station. Each bus line follows a fixed route and operates on a set daily schedule. Each route forms a loop with multiple stops, beginning and ending at the same terminal station. The set of trips for all bus lines in a day is indexed by $k\in \{1,2,\ldots, K\}$. Let $T^{\rm d}_{k}$ denote the departure time step for the $k$-th trip of the day in chronological order. \par

The bus lines are served by a set of EBs, indexed by $i\in\mathcal{M}=\{1,2,\ldots, M\}$, where each EB $i$ is assigned to at most one trip $k$ at any given time step. The trip assigned to EB $i$ at time step $t$ is denoted by $k_{i,t}\in \{0,1,\ldots,K\}$, where $k_{i,t}=0$ indicates that the EB is not assigned to any trip.

There are two alternating periods for each EB, as shown in Fig. \ref{systemmodel}. One is the layover period, in which the EB stays at the terminal station. The other is the operating period, in which the EB travels along the route of the assigned trip. We use $B_{i,t}$ to denote the operating status of EB $i$, where $B_{i,t}=1$ and $B_{i,t}=0$ correspond to the layover and operating periods, respectively. \par 

Any EB $i$ in the layover period must depart the terminal station and enter the operating period at the scheduled departure time of its currently assigned trip, i.e., EB $i$ switches from the layover period to the operating period at time step $t$ if $T^{\rm d}_{k}=t$ and $k_{i,t}=k$. When EB $i$ completes the trip $k$ and arrives at the terminal station at time step $t'$, it switches from the operating period to the layover period. A new trip $k'=k_{i,t'}$ is assigned to EB $i$ upon its arrival at the terminal station. The assigned trip $k'$ for EB $i$ can be changed at any time step during its current layover period. \par

Let $\varGamma_{i,t}$ denote the probability that the current layover or operating period for EB $i$ is terminated at time step $t$. 
To derive this probability, we first define a variable $\tau_{i,t}$ as 
\begin{equation}
	\label{taut}
	\\ \tau _{i,t}=\begin{cases}  T^\mathrm{d}_{k_{i,t} }-t-1,&		\mathrm{if}\,\,B_{i,t}=1 \\ \tau _{i,t-1}+1, &	\mathrm{if}\,\,B_{i,t}=0 \\\end{cases},
\end{equation}
\noindent where $T^\mathrm{d}_{k_{i,t} }$ is the departure time step for the trip $k_{i,t}$ assigned to EB$i$ at time step $t$. Let $T^\mathrm{d}_{k_{i,t}}=\infty$ if $k_{i,t}=0$. During the layover period when $B_{i,t}=1$, $\tau_{i,t}$ represents the remaining number of time steps from time step $t$ until the departure time $T^\mathrm{d}_{k_{i,t}}$ of trip $k_{i,t}$ currently assigned to EB $i$. In contrast, during the operating periods when $B_{i,t}=0$, $\tau_{i,t}$ denotes the number of time steps from the last departure time of EB $i$ to the current time step $t$.

	

Note that $\varGamma_{i,t}$ depends on both $B_{i,t}$ and $\tau_{i,t}$. We consider that the travel time per trip is random due to the uncertain traffic conditions. During the operating period when $B_{i,t}=0$, the probability of EB $i$ returning to the terminal station increases as its traveling time elapses. Let $T^{\rm o}_i$ be a random variable that represents the duration of an operating period for EB $i$. The termination probability for the operating period can be expressed as
\begin{align}
	\label{gammaBt=0}
	\varGamma_{i,t}(B_{i,t}=0,\tau_{i,t})={\rm Pr}(B_{i,t+1}=1|B_{i,t}=0,\tau_{i,t}) \IEEEnonumber \\
 =\frac{\mathrm{Pr}(T^{\rm o}_i=\tau_{i,t})}{\prod_{x=0}^{\tau_{i,t}-1}(1-\mathrm{Pr}(T^{\rm o}_i=x))}, 
\end{align}
\noindent \noindent where the numerator represents the probability of the EB $i$ arriving at the terminal station at time step $t$, while the denominator represents the probability of the EB $i$ not arriving at the terminal station before time step $t$.

The layover period with $B_{i,t}=1$ terminates with probability $1$ when $\tau _{i,t}=0$, indicating the scheduled departure time for EB $i$ is reached at the next time step. The termination probability for the layover period can thus be expressed as
\begin{align}
	\label{gammaBt=1}
 \varGamma_{i,t}(B_{i,t}=1,\tau_{i,t})&={\rm Pr}(B_{i,t+1}=0|B_{i,t}=1,\tau_{i,t}) \IEEEnonumber \\ 
 &=\begin{cases}  1,&		\mathrm{if}\,\,\tau _{i,t}=0 	\\ 0,&		\mathrm{if}\,\,\tau _{i,t}>0 \\\end{cases}.
\end{align} 

\subsection{EB charging model}

We consider $N$ chargers in the terminal station, where the number of chargers is smaller than that of EBs, i.e., $N<M$. We use $\mathcal{M}^{\rm lay}_t$ to denote the set of EBs that are currently in the layover period at time step $t$, i.e., $\mathcal{M}^{\rm lay}_t=\{i|i\in\mathcal{M},B_{i,t}=1\}$. Let $M_{t}=|\mathcal{M}^{\rm lay}_t|=\sum_{i=1}^M{B_{i,t}}$ denote the number of EBs in the layover period. If $M_{t}>N$, only $N$ EBs can be charged, and the rest of the $M_{t}-N$ EBs have to enter the waiting area. For the sake of simplicity, we assume that the time to switch an EB from a charger to the waiting area and vice versa is negligible \cite{ye2022learning}.\par 

For each EB $i$, let $\omega_{i,t}\in\{0,1\}$ denote the charging status at time step $t$, where $\omega_{i,t}=1$ stands for charging, and $\omega_{i,t}=0$ stands for not charging. Only the EBs in the layover period can be allocated a charger, i.e., $\omega_{i,t}=1$. When $\omega_{i,t}=0$, the EB is either waiting to be charged if $B_{i,t}=1$ or operating if $B_{i,t}=0$. Due to the limited number of chargers, $\omega_{i,t}$ is constrained by
\begin{equation}
\label{omega_constraint}
\sum_{i=1}^M{\omega _{i,t}\leqslant N} .
\end{equation} 

Let $E_{i,t}$ denote the State of Charge (SoC) level of the battery for the EB $i$ at time step $t$, which is constrained by the maximum and minimum battery capacity $E_{\max}$ and $E_{\min}$, i.e., $E_{\min}\leqslant E_{i,t}\leqslant E_{\max}$.

Let $c_{i,t}$ denote the charging power of EB $i$ at time step $t$. When $B_{i,t}=1$ and $\omega_{i,t}=1$, the EB is allocated a charger so it can either charge energy from or discharge energy back to the electric grid in the V2G mode. When $B_{i,t}=1$ and $\omega_{i,t}=0$, the EB is waiting so the charging power is zero, i.e., $c_{i,t}=0$. When $B_{i,t}=0$, the EB is continuously discharging since it travels along the bus route, resulting in a negative value for $c_{i,t}$. Thus, the space of $c_{i,t}$, denoted as $\mathcal{C}_{i,t}$, is derived as 
\begin{equation}
	\label{ct_constraint}
	\mathcal{C}_{i,t} = \begin{cases}
		\left[ -d_{\max},c_{\max} \right]\cap  &	\\
               \left[ \frac{E_{\min}-E_{i,t}}{\varDelta t},\frac{E_{\max}-E_{i,t}}{\varDelta t} \right], & \mathrm{if}\, B_{i,t}=1 \ \& \ \omega_{i,t}=1  \\
		\{0\}, & \mathrm{if}\,B_{i,t}=1 \ \& \ \omega_{i,t}=0\\
		\left[ -d_{\max}, 0 \right]\cap \left[ \frac{E_{\min}-E_{i,t}}{\varDelta t},0 \right] ,&		\mathrm{if}\,B_{i,t}=0\\
	\end{cases},
\end{equation}

\noindent where $c_{\max}$ and $d_{\max}$ denote the maximum absolute value of charging and discharging power, respectively. In addition, $\left[ \left( E_{\min}-E_{i,t} \right) /\varDelta t,\left( E_{\max}-E_{i,t} \right) /\varDelta t \right]$ represents the value range due to the limitation of EB battery capacity.

Finally, the dynamics of the EB battery can be modeled as
\begin{equation}
	\label{Etdynamic}
 E_{i,{t+1}}=E_{i,t}+ c_{i,t}\cdot \Delta t.
\end{equation}

\subsection{Trip Assignment model}
At each time step $t$, a trip is assigned to each EB $i\in\mathcal{M}^{\rm lay}_t$ in the layover period. Let $\hat{k}_t$ denote the earliest upcoming trip departing after time step $t$, defined as 
\begin{equation}
\label{hatk}
\hat{k}_t={\mathrm{arg}\underset{k}{\min}}T_{k}^{\mathrm{d}}, \forall T_{k}^{\mathrm{d}}>t.
\end{equation}

We define a permutation of EBs in the layover period as $P(\mathcal{M}_t^{\rm lay})=(j_1,j_2,\ldots,j_m,\ldots,j_{M_t})$, where $j_m \in \mathcal{M}_t^{\rm lay}$ and each $j_m$ is unique for $m \in \{1,\ldots,M_t\}$. In this ordering, the earlier an EB's index appears in $P(\mathcal{M}_t^{\rm lay})$, the sooner it's assigned trip will depart. Thus, for any two elements $j_m$ and $j_{m'}$ in $P(\mathcal{M}_t^{\rm lay})$ with $m<m'$, we have $k_{j_m}<k_{j_{m'}}$. Consequently, given the permutation $P(\mathcal{M}_t^{\rm lay})$, the trip $k_{j_m,t}$ of each EB $j_m\in \mathcal{M}_t^{\rm lay}$ can be determined by
\begin{align}
\label{trip_map}
k_{j_m,t}=
\begin{cases}
	\hat{k}_t+m-1&		\mathrm{if} \ \hat{k}_t+m-1\leq K\\
	0 &		\mathrm{if} \ \hat{k}_t+m-1 > K\\
\end{cases}, \\ \IEEEnonumber
 \forall m\in \{1,\ldots,M_t\}.
\end{align}
\noindent The second condition in \eqref{trip_map} applies when the number of future trips to depart is fewer than the number of EBs in the layover period, i.e., $K-\hat{k}_t+1<M_t$. In this case, the last $\hat{k}_t+M_t-1-K$ EBs are not assigned any trip, resulting in $k_{j_m,t}=0$. \par

Based on \eqref{trip_map}, we can define a mapping function 
\begin{equation}
    f: P(\mathcal{M}_t^{\rm lay}) \rightarrow\{k_{i,t}\}_{i\in\mathcal{M}^{\rm lay}_t}
    \label{trip function}
\end{equation} 
\noindent from a permutation of EBs in the layover period to the corresponding trip assignment for these EBs. \par

\subsection{Optimization objective}

The objective function is defined as 
\begin{align}
    \label{optimization function}
    \mathrm{Minimize}\,\,\mathbb{E} \left[ \sum_{t=0}^{T-1}{\left( \sum_{i=1}^M{(C_{i,t}^{\mathrm{ch}}+C_{i,t}^{\mathrm{ba}}+C_{i,t}^{\mathrm{sw}})}+C_{t}^{\mathrm{end}} \right)} \right] ,
\end{align}
\noindent where $C_{i,t}^{\mathrm{ch}}$, $C_{i,t}^{\mathrm{ba}}$, $C_{i,t}^{\mathrm{sw}}$, and $C_{t}^{\mathrm{end}}$ represent different operational costs. Specifically, $C_{i,t}^{\mathrm{ch}}$ is the charging cost derived as 
	\begin{equation}
		\label{chargingcost}
		C^{\mathrm{ch}}_{i,t}={\rho_t c_{i,t}B_{i,t}\varDelta t},
	\end{equation}

\noindent where $\rho_t$ denotes the electricity price at time step $t$. 

$C_{i,t}^{\mathrm{ba}}$ denotes the battery degradation cost, which is positively correlated with the absolute value of the charging power $c_{i,t}$:
	\begin{equation}
		\label{batterycost}
		C_{i,t}^{\mathrm{ba}}={C^\mathrm{b}\left| \frac{b_k}{100} \right|\left| \frac{c_{i,t}}{E_{\max}} \right|B_{i,t}},
	\end{equation}
\noindent where $C^\mathrm{b}$ is a constant representing the total battery degradation cost, and $b_k$ represents the slope of the linear approximation of the battery life as a function of the cycles \cite{ortega2014optimal}. \par

Next, $C_{i,t}^{\mathrm{sw}}$ is the switching cost to denote the penalty of frequently switching the charging status of the EBs that are in the layover periods, i.e.,
\begin{equation}
	\label{changingcost}
	C_{i,t}^{\mathrm{sw}}={C^\mathrm{s}\omega _{i,t-1}\left( 1-\omega _{i,t} \right)B_{i,t}},
\end{equation}
\noindent where $C^\mathrm{s}$ is a constant value. 


Finally, $C_t^{\mathrm{end}}$ represents the penalty incurred when the SoC level in any EB's battery falls below the minimum battery capacity $E_{\min}$ during operation:
\begin{equation}
\label{Costend}
C_{t}^{\mathrm{end}}=\begin{cases}
	C^{\mathrm{E}}&		\exists i\in \mathcal{M} ,E_{i,t}<E_{\min}\\
	0&		\mathrm{otherwise}\\
\end{cases}.
\end{equation}
\noindent Here, $C^{\mathrm{E}}$ is a large constant to strongly discourage any EB from running out of battery during operation. \par

There are three decision variables, i.e., the charging power decision $c_{i,t}$, the charger allocation decision $\omega_{i,t}$, and the trip allocation decision $k_{i,t}$. The constraints for $c_{i,t}$, $\omega_{i,t}$, and $k_{i,t}$ are given by \eqref{ct_constraint}, \eqref{omega_constraint}, and \eqref{trip_map}, respectively.

\section{MDP Model}
\subsection{Original MDP}

\subsubsection{State} The global state $S_t$ aggregates the local states $S_{i,t}$ for each EB $i$, such that $S_t=\{S_{i,t}\}_{i=1}^M$, where $S_{i,t}=\left\lbrace E_{i,t},B_{i,t},B_{i,t-1},\tau_{i,t-1},\omega_{i,{t-1}},H_t,t\right\rbrace$. Without loss of generality, let $\omega_{i,{-1}}=0$. Here, $H_t$ denote the historical electricity prices from time step $t-h$ up to $t$, i.e., $H_t=\left( \rho_{t-h}, \rho_{t-h+1},\ldots, \rho_{t-1}, \rho_t\right)$,
\noindent where $h$ is the length of the time window for past prices. Since $H_t$ and $t$ are common across all $S_{i,t}$, only one instance of $H_t$ and $t$ is included in $S_t$ after aggregation. 

Let $\mathcal{S}^{+}$ denote the state space, which can be divided into the set of nonterminal states $\mathcal{S}=\{S_{t}|E_{i,t}\geq E_{\min}, \forall i \in \mathcal{M}\}$ and the set of terminal states $\mathcal{S}^{\mathrm{T}}=\{S_{t}|E_{i,t}< E_{\min}, \exists i \in \mathcal{M}\}$. This means that when the SoC level in any EB's battery is lower than the minimal battery capacity constraint, i.e., $E_{i,t}< E_{\min}$, the agent will enter the terminal states and the current episode will end before the maximum time step $T$ is reached. 

\subsubsection{Action} 
At each time step $t$, the agent only determines the actions of those EBs that are currently in the layover period ($i\in\mathcal{M}^{\rm lay}_t$). Let $A_{i,t}=\{c_{i,t}, \omega_{i,t}, k_{i,t}\}$ represent the local action for EB $i$ at time step $t$. The global action $A_t$ is the aggregation of the local actions $A_{i,t}$ for all EBs in the layover period, i.e., $A_t=\{A_{i,t}\}_{i\in\mathcal{M}^{\rm lay}_t}$. Correspondingly, $A_t=(c_t,\omega_t,k_t)$ consists of three components: the charging power action, denoted by $c_t=\{c_{i,t}\}_{i\in\mathcal{M}^{\rm lay}_t}$, the charger allocation action, denoted by $\omega_{t}=\{\omega_{i,t}\}_{i\in\mathcal{M}^{\rm lay}_t}$, and the trip assignment action, denoted by $k_{t}=\{k_{i,t}\}_{i\in\mathcal{M}^{\rm lay}_t}$.  \par

Let $\mathcal{A}_t$ represent the state-dependent action space at time step $t$, where $\mathcal{A}_t=\mathcal{C}_t\times\Omega_t\times\mathcal{K}_t$. 
The charging power action space can be expressed as $\mathcal{C}_t=\Pi_{i\in \mathcal{M}^{\rm lay}_t}\mathcal{C}_{i,t}$, where $\mathcal{C}_{i,t}$ is given by \eqref{ct_constraint}. The charger allocation action space $\Omega_t$ is defined as
\begin{equation}
\label{omega_space}
\Omega_t =\left\{ \{\omega _{i,t}\}_{i\in\mathcal{M}^{\rm lay}_t}\bigg |\omega _{i,t}\in \left\{ 0,1 \right\} ,\sum_{i=1}^M{\omega _{i,t}\leqslant N} \right\},
\end{equation} \par
\noindent where the last constraint is due to the limited number of chargers. \par

The trip assignment action space $\mathcal{K}_t$ is defined as 
\begin{align}
  \label{linespace}
      \mathcal{K}_t =& \Bigg\{ \{k_{i,t}\}_{i\in\mathcal{M}^{\rm lay}_t} \bigg |
f: P(\mathcal{M}_t^{\rm lay}) \rightarrow\{k_{i,t}\}_{i\in\mathcal{M}^{\rm lay}_t}, \forall P(\mathcal{M}_t^{\rm lay})   \Bigg\},
\end{align}

\noindent where the mapping function $f$ is provided in \eqref{trip function}. \par


When an EB $i$ is in the operating period ($B_{i,t}=0$), the action $A_{i,t}$ is given by the environment rather than being determined by the agent. Specifically, the charging power action $c_{i,t}$ is a random variable whose range is specified by the third case in \eqref{ct_constraint}. The charger allocation action $\omega_{i,t}$ is always zero. The trip assignment action $k_{i,t}$ remains the same as that at the previous step $t-1$, i.e., $k_{i,t}=k_{i,t-1}$. 

\subsubsection{Transition Probability}
 	The state transition probability is derived as
 	\begin{align}
 			\mathrm{Pr}\left( S_{t+1}|S_t,A_t \right) =\mathrm{Pr}\left( H_{t+1}|H_t \right)\mathrm{Pr}\left( t+1|t \right) \IEEEnonumber \\
 			\prod_{i=1}^M{\bigg[ \mathrm{Pr}\left( E_{i,t+1}|E_{i,t},c_{i,t} \right)}\mathrm{Pr}\left( B_{i,t+1}|B_{i,t},\tau _{i,t} \right)   \IEEEnonumber \\
 		\mathbf{1}_{B_{i,t}}\,\,\mathbf{1}_{\omega _{i,t}}	\,\,\mathrm{Pr}\left( \tau _{i,t}|\tau _{i,t-1},B_{i,t},k_{i,t},t \right)\bigg],
 		\label{pr}
 	\end{align}
 \noindent where the transition probabilities of historical electricity prices $\mathrm{Pr}\left( H_{t+1}|H_{t} \right)$ is not available, but samples of the trajectory can be obtained from real-world data. The transition probability of time steps is always $\mathrm{Pr}\left( t+1|t \right)=1$. The transition probability of SoC level for each EB, i.e., $\mathrm{Pr}\left( E_{i,t+1}|E_{i,t},c_{i,t} \right)$ can be derived from \eqref{Etdynamic}. Next, the transition probability $\mathrm{Pr}\left( B_{i,t+1}|B_{i,t},\tau _{i,t} \right)$ can be derived from the termination probability $\varGamma_{i,t}(B_{i,t},\tau _{i,t})$, i.e.,
 	\begin{align}
 		\label{prtaut}
 		&\mathrm{Pr}\left( B_{i,t+1}|B_{i,t},\tau _{i,t} \right)= \\ \IEEEnonumber &\begin{cases}	1-\varGamma_{i,t}(B_{i,t},\tau _{i,t}) ,&		\mathrm{if}\,\,B_{i,t+1}= B_{i,t}\\	\varGamma_{i,t}(B_{i,t},\tau _{i,t}),&		\mathrm{if}\,\,B_{i,t+1}= 1-B_{i,t} \\\end{cases},
 	\end{align}
 	\noindent where $\varGamma_{i,t}(B_{i,t},\tau _{i,t})$ is given by \eqref{gammaBt=1} and \eqref{gammaBt=0}. Finally, the transition probability $\mathrm{Pr}\left( \tau _{i,t}|\tau _{i,t-1},B_{i,t},k_{i,t},t \right)$ for $\tau_{i,t}$ is derived from \eqref{taut}. \par
 	 
\subsubsection{Reward function} 
The objective of a MDP model is to derive the optimal policy $\pi^*$ that maximizes the expected return, where the return is defined as the sum of rewards: 
\begin{equation}
	\label{objective1}
	\pi ^*=\mathrm{arg}\underset{\pi}{\max}\mathbb{E}\left[ \sum_{t=0}^{T-1}{r\left( S_t,A_t \right)} \right]. 
\end{equation}

Since \eqref{objective1} must align with the optimization objective in \eqref{optimization function} in Section III.D, we can derive the reward function $r\left( S_{t},A_{t} \right)$ as
\begin{equation}
	\label{rewardEB}
 r\left( S_{t},A_{t} \right) = \sum_{i=1}^M{r_i\left( S_{i,t},A_{i,t} \right)}-C_t^{\mathrm{end}},
\end{equation} 
\noindent where the reward of each EB $r_i\left( S_{i,t},A_{i,t} \right)$ is
	\begin{equation}
		\label{rewardEBi}
		r_i\left( S_{i,t},A_{i,t} \right) =-C_{i,t}^{\mathrm{ch}}-C_{i,t}^{\mathrm{ba}}-C_{i,t}^{\mathrm{sw}}.
	\end{equation}

 Since $C_{i,t}^{\mathrm{sw}}$ depends on $\omega_{i,{t-1}}$ according to \eqref{changingcost}, we include $\omega_{i,{t-1}}$ in the state $S_{i,t}$ defined in Section IV.A.1). \par 
	 


\subsection{Options over MDP}   
The original MDP model defined in Section III.A involves two types of actions that can operate at different time scales. Specifically, a course of both charger and trip allocation actions can persist for a variable period of time, while the charging power actions are taken per time step. In order to take advantage of the simplicities and efficiencies of temporal abstraction, we adopt the framework of options to abstract actions at two temporal levels. Let $o_t=\{\omega_t,k_t\} \in \mathcal{O}_t $ denote the options, where $\mathcal{O}_t=\varOmega_t\times\mathcal{K}_t$ is the state-dependent option space.\par
Options can be regarded as temporally extended ``actions", which can last for multiple time steps \cite{sutton1999between}. An option is prescribed by the \emph{policy over options} $\mu:\mathcal{S}\times\mathcal{O}_t\rightarrow[0,1]$, where an option $o_t \in \mathcal{O}_t $ is selected according to the probability distribution $\mu(o_t|S_{t})$. Each option $o\in \mathcal{O}_t$ is associated with a triple, i.e., $\left( \mathcal{I}_{o} , \pi_{o} , \beta_{o} \right)$. $\mathcal{I}_{o}\subseteq\mathcal{S}$ is the initiation set of states, i.e., $o$ is available in state $S_t$ if and only if $S_t \in \mathcal{I}_{o}$. $\pi_{o}:\mathcal{S}\times\mathcal{C}_t\rightarrow[0,1]$ is the \emph{intra-option policy} and $\beta_{o}:\mathcal{S}^{+}\rightarrow[0,1]$ is the termination condition. Considering that only the EBs in the terminal station can be charged at each time step, the initiation set $\mathcal{I}_{o}$ is defined as 
\begin{equation}
	\label{initI}
	\mathcal{I}_{o}=\left\{ S_t|B_{i,t}=1,\exists i \in \mathcal{M}\right\} .
\end{equation}

The policy over options $\mu$, the intra-option policy $\pi_{o}$, and the termination condition $\beta_{o}$ for each option $o\in\mathcal{O}_t$ all have to be learned. Note that the current option is forced to terminate when an EB departs or returns to the terminal station, as changes in $\mathcal{M}_t^{\rm lay}$ lead to a corresponding change in the action space $\mathcal{A}_t$. Therefore, we have
\begin{equation}
\label{betafunction}
\beta_{o_{t-1}}(S_t)=
1, \ \mathrm{if}\, B_{i,t-1}+B_{i,t}=1. 
\end{equation}
\noindent Since $B_{i,t-1}$ is required to calculate \eqref{betafunction}, it is included in the state $S_{i,t}$ defined in Section IV.A.1).
\subsection{Two Augmented MDPs}
In order to efficiently learn $\mu$ as well as $\pi_{o}$ and $\beta_{o}$, $\forall o\in\mathcal{O}_t$, we reformulate the options over MDP as two augmented MDPs, i.e., the high-level MDP $\mathcal{M}^\mathrm{H}$ and the low-level MDP $\mathcal{M}^\mathrm{L}$, based on the DAC architecture. The optimal policy over options $\mu^{*}$ and the optimal termination condition $\beta_{o}^{*}$ for each option $o\in\mathcal{O}_t$ can be derived by solving $\mathcal{M}^\mathrm{H}$, while the optimal intra-option policy $\pi_{o}^{*}$ for each option $o\in\mathcal{O}_t$ can be derived by solving $\mathcal{M}^\mathrm{L}$. \par

\newtheorem{mydef}{Definition}
\begin{mydef}[High-Level MDP]
	Given the intra-option policy $\pi_{o}$ for $o\in\mathcal{O}_t$, we define the high-level MDP $\mathcal{M}^\mathrm{H}$ as follows:
	\begin{itemize}
		\item State $S^\mathrm{H}_t=(S_t,o_{t-1})=(S_t,k_{t-1})$. To maintain the Markov property, $S_t$ in the original MDP should be augmented by including the option of the previous time step $o_{t-1}$ \cite{zhang2019dac}. Since $\omega_{t-1}$ is already an element of $S_t$ by its definition, the state space of $\mathcal{M}^\mathrm{H}$ is $\mathcal{S}^\mathrm{H}=\mathcal{S}^{+}\times\mathcal{K}_{t-1}$. Without loss of generality, let $k_{i,-1}=0$ for all $i\in \mathcal{M}$. 
		\item Action $A^\mathrm{H}_t=o_t=(\omega_t,k_t)$. The action in $\mathcal{M}^\mathrm{H}$ is also the option. Therefore, the action space of $\mathcal{M}^\mathrm{H}$ is $\mathcal{O}_t$.
		\item The reward function $r^{\mathrm{H}}(S_t^{\mathrm{H}},A_t ^{\mathrm{H}})=r^{\mathrm{H}}(S_t,k_{t-1},\omega_t,k_t)=\sum_{c\in\mathcal{C}_t}\pi_{o_t}(c|S_{t})r(S_t,\omega_t,k_t,c)$, where $r(S_t,\omega_t,k_t,c)$ is the reward function of the original MDP defined in \eqref{rewardEB}. 
	\end{itemize}
\end{mydef}

The high-level policy $\pi^\mathrm{H}$ of $\mathcal{M} ^{\mathrm{H}}$ is defined as 
\begin{align}
\label{policy mu}
\pi^\mathrm{H} \left( A_{t}^{\mathrm{H}}|S_{t}^{\mathrm{H}} \right) = \pi^\mathrm{H} \left(o_t|S_t,o_{t-1}\right)=\mathrm{Pr}\left( o_t|S_t,o_{t-1} \right) \IEEEnonumber \\ 
=\left( 1-\beta_{o_{t-1}}(S_t) \right) \mathbb{I}_{o_t=o_{t-1}} +\beta _{o_{t-1}}(S_t)\mu \left(o_t| S_t\right) ,
\end{align}
\noindent where $\mathbb{I}$ is the indicator function. Note that $\pi^\mathrm{H}$ is a composite function of the policy over options $\mu$ and the termination condition $\beta_{o_{t-1}}$. Therefore, \eqref{policy mu} implies that with probability $\left( 1-\beta_{o_{t-1}}(S_t) \right)$ the option will remain unchanged, i.e., $o_t=o_{t-1}$, but with probability $\beta_{o_{t-1}}(S_t)$ it will terminate. When the option terminates, the policy $\mu \left(o_t| S_t\right)$ is used to generate a new option.

The transition probability of $\mathcal{M} ^{\mathrm{H}}$ is defined as
 	\begin{align}
 			p^{\rm H}\left( S_{t+1}^{\rm H}|S^{\rm H}_t,A^{\rm H}_t \right) \doteq\mathrm{Pr}\left( S_{t+1},o_t|S_t,o_{t-1},o_t \right) \IEEEnonumber \\
        =\mathrm{Pr}\left( S_{t+1}|S_t,o_t \right)=\sum_{c\in\mathcal{C}_t}\pi_{o_t}(c|S_{t})\mathrm{Pr}\left( S_{t+1}|S_t,c,o_t \right),
 		\label{pr_high}
 	\end{align}
\noindent where $\mathrm{Pr}\left( S_{t+1}|S_t,c,o_t \right)$ is given in \eqref{pr}.

\begin{mydef}[Low-Level MDP]
	Given the policy over options $\mu$ and the termination condition $\beta_{o}$ for $o\in\mathcal{O}_t$, we define the low-level MDP $\mathcal{M}^\mathrm{L}$ as follows:
	\begin{itemize}
		\item State $S^\mathrm{L}_t=(S_t,o_t)=(S_t,\omega_t,k_t)$. The state $S_t$ in the original MDP is augmented by including the option of the current time step $o_{t}$. Thus, the state space of $\mathcal{M}^\mathrm{L}$ is $\mathcal{S}^\mathrm{L}=\mathcal{S}^{+}\times \mathcal{O}_t$. 
		\item Action $A^\mathrm{L}_t=c_t$. The action in $\mathcal{M}^\mathrm{L}$ is the charging power action in the original MDP. Therefore, the action space of $\mathcal{M}^\mathrm{L}$ is $\mathcal{C}_t$.
		\item The reward function $r^{\mathrm{L}}(S_t^{\mathrm{L}},A_t ^{\mathrm{L}})=r(S_t,\omega_t,k_t,c_t)=r(S_t,A_t)$, which is the reward function of the original MDP defined in \eqref{rewardEB}.
	\end{itemize}
\end{mydef}
The low-level policy $\pi^\mathrm{L}$ of $\mathcal{M} ^{\mathrm{L}}$ is defined as 
\begin{align}
	\label{policy pi L}
	&\pi^\mathrm{L} \left( A_{t}^{\mathrm{L}}|S_{t}^{\mathrm{L}} \right) =	\mathrm{Pr}\left( c_t|S_t,o_t \right) =\pi_{o_{t}}(c_t|S_t),
\end{align}
\noindent where $\pi_{o_{t}}$ is the intra-option policy of $o_{t}$. \par 

The transition probability of $\mathcal{M} ^{\mathrm{L}}$ is defined as
 	\begin{align}
 			p^{\rm L}\left( S_{t+1}^{\rm L}|S^{\rm L}_t,A^{\rm L}_t \right) \doteq \mathrm{Pr}\left( S_{t+1},o_{t+1}|S_t,o_{t},c_t \right) \IEEEnonumber \\
    =\mathrm{Pr}\left( S_{t+1}|S_t,A_t \right)\mathrm{Pr}\left( o_{t+1}|S_{t+1},o_t \right),
 		\label{pr_low}
 	\end{align}
    \noindent where $\mathrm{Pr}\left( S_{t+1}|S_t,A_t \right)$ is given in \eqref{pr} and $\mathrm{Pr}\left( o_{t+1}|S_{t+1},o_t \right)$ is given in \eqref{policy mu}.
    
\section{HDRL solution}
\subsection{The Basic DAC-MAPPO Algorithm}
\begin{algorithm}[t]
	\caption{The basic DAC-MAPPO algorithm}
	\label{alg-DAC-MAPPO}
	\begin{algorithmic}[1]  
		\State Randomly initialize the high-level actor network ${\mu}_\theta(o_t|S_t)$ with parameter $\theta$, the low-level actor network ${\pi}_\vartheta(c_{i,t}|S_{i,t},o_t)$ with parameter $\vartheta$, the critic network ${V}_\phi(S_t,o_t)$ with parameter $\phi$, and the terminal condition network $\beta_{o_{t-1},\varphi}(S_t)$ with parameter $\varphi$. 
		\For{episode $e = 1, ..., E$}
		\State Initialize the start state $S_0$, $S_t \gets S_0$
            \State Initialize the terminal condition $\beta_{o_{t-1}}(S_t) \gets 1$
		\For{$t = 0, ..., T-1$}
		\State Calculate the high-level policy $\pi^{\rm H}$ based on the terminal condition $\beta_{o_{t-1}}(S_t) $ and the network ${\mu}_\theta(o_t|S_t)$ according to \eqref{policy mu}
		\State Sample an option $o_{t}$ from the high-level policy $\pi^{\rm H}$
		\For{EB $i = 1, ..., M$ }
		\If{$\omega_{i,t}=1$}
		\State Sample an action $c_{i,t}$ from the network ${\pi}_\vartheta(c_{i,t}|S_{i,t},o_t)$ 
		\Else
		\State Observe $c_{i,t}$ from environment
		\EndIf
  \EndFor
  \State Execute the global action $c_{t}=\{c_{i,t}\}_{i=1}^{M}$ and observe reward $r_{t+1}$, and next state $S_{t+1}$ from environment
  \State Update the terminal condition by the network $\beta_{o_t}(S_{t+1})=\beta_{o_{t},\varphi}(S_{t+1})$
		\EndFor
  \State Optimize $\theta$, $\phi$, and $\varphi$ based on PPO 
  \State Optimize $\vartheta$ and $\phi$ based on MAPPO 
		\EndFor
	\end{algorithmic}  
\end{algorithm} 
In this section, we propose a basic DAC-MAPPO algorithm to solve the two augmented MDPs presented in Section IV.C and learn both the optimal high-level and low-level policies. Building upon the DAC architecture, this algorithm incorporates a two-level hierarchical framework, which will be described in detail separately. 
At the high level, since the high-level policy needs to be determined based on the global state $S_t$, we employ a single centralized high-level agent that observes $S_t$. The high-level augmented MDP is solved using the Proximal Policy Optimization (PPO) algorithm to learn the high-level policy $\pi^{\rm H}$. As $\pi^{\rm H}$ is a compound policy that integrates both $\mu(o_t|S_t)$ and $\beta_{o_{t-1}}(S_t)$, as defined in \eqref{policy mu}, the high-level agent uses two distinct networks: the high-level actor network ${\mu}_\theta(o_t|S_t)$ with parameter $\theta$ and the terminal condition network $\beta_{o_{t-1},\varphi}(S_t)$ with parameter $\varphi$, to approximate $\mu(o_t|S_t)$ and $\beta_{o_{t-1}}(S_t)$, respectively. It is important to note that $\beta_{o_{t-1},\varphi}(S_t)$ plays a crucial role by determining the termination of options, thereby effectively preventing any EB from occupying a charger for an extended period.

At the low level, we employ decentralized agents, treating each EB as an agent that observes its local augmented state, $(S_{i,t},o_t)$. This approach can not only enhance scalability as the number of EBs increases but also achieve faster convergence by reducing the dimensionality of the state space. Since the termination condition learned at the high level influences the charging duration for the allocated EBs at the low level, the low-level charging power decisions of each EB are not entirely independent, which leads to inevitable interactions among the EBs. To effectively manage the multi-agent problem arising from these interactions, we utilize the MAPPO algorithm, rather than independent PPO, to solve the low-level augmented MDP.\par
In MAPPO, the low-level local policy $\pi_i^{\rm L}\left( c_{i,t}|S_{i,t},o_t \right)$ for an EB $i$ is derived using a decentralized actor network. Parameter sharing technique is applied, where all low-level agents share the same actor network ${\pi}_\vartheta(c_{i,t}|S_{i,t},o_t)$ with parameter $\vartheta$ to expedite the training. Moreover, a centralized critic network ${V}_\phi(S_t,o_t)$ with parameter $\phi$ is used at the low level to approximate the low-level value function by ${V}^{\rm L}(S_t^{\rm L})\approx {V}_\phi\left( S_t,o _t \right)$. Since the high-level value function can be derived from the low-level value function according to ${V}^{\rm H}(S_t^{\rm H})=\sum_{o_t\in \mathcal{O}_t}^{}{\pi ^{\mathrm{H}}\left( o_t|S_t \right) {V}^{\rm L}(S_t^{\rm L})}$, only one critic ${V}_\phi(S_t,o_t)$ is needed for both levels \cite{zhang2019dac}.\par

The pseudocode of DAC-MAPPO is detailed in Algorithm \ref{alg-DAC-MAPPO}. Note that the parameter $\phi$ of the critic is updated twice per iteration - once by PPO and once by MAPPO - since a single critic network is shared between the high-level and low-level policies. For further technical details, the optimization for PPO and MAPPO is referenced in \cite{chen2018adaptive} and \cite{yu2022surprising}, respectively.

\subsection{The Enhanced DAC-MAPPO-E algorithm with Improved Scalability}

\subsubsection{High-Level Actor Architecture for Large Action Space}

\begin{figure*}[ht]
\centering
\includegraphics[width=0.7\linewidth]{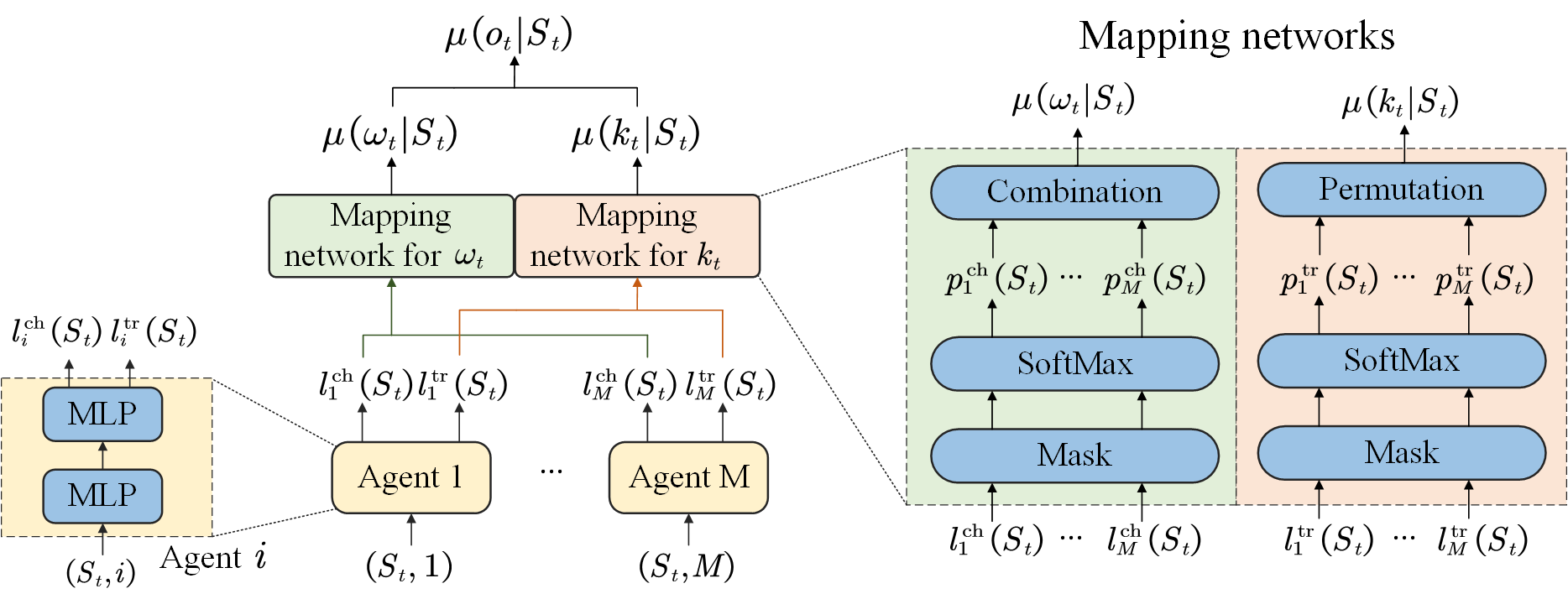}
\caption{The new decentralized high-level actor network is designed by decoupling the high-level action space, with an architecture comprising $M$ agent networks and a pair of mapping networks. Each agent network is associated with an EB. The mapping networks are utilized to derive the policy over options $\mu(o_t|S_t)$.}
\label{new-high-level-actor} 
\end{figure*}

The high-level actor in DAC-MAPPO is a function approximator that scales linearly with the number of options in the option space $\mathcal{O}_t$. The option space comprises two subspaces: $\varOmega_t$ and $\mathcal{K}_t$, both of which grow substantially as the number of EBs increases. Specifically, the size of $\varOmega_t$ corresponds to the number of ways to select $N$ EBs from a total of $M$ EBs, given by $\left( \begin{array}{c}M\\N\\ \end{array} \right)  = \frac{M!}{N!(M-N)!}$. Meanwhile, the maximum size of $\mathcal{K}_t$ is the number of permutations of all $M$ EBs, i.e., $M!$. This results in the high-level action space being of size $\frac{M!}{N!(M-N)!} \times M!$, which grows super-exponentially with $M$ due to the factorial term. This super-exponential growth in the action space as the number of EBs increases leads to an extremely large number of parameters in the actor's output layer and high computational complexity of action sampling. \par 

One potential solution to address this challenge is directly adopting decentralized high-level actors that are similar to the approach used for low-level actors. However, this is infeasible due to the mutual exclusion of individual high-level actions per EB. Specifically, a single charger or a trip cannot be allocated to more than one EB simultaneously. To address this complexity, we design the high-level actor network with an architecture comprising $M$ agent networks and a pair of mapping networks. \par

As illustrated in Fig. \ref{new-high-level-actor}, an agent network is associated with each EB $i\in \mathcal{M}$. The network takes the global state $S_t$ and the index $i$ as input, and outputs a pair of logits: $l^{\rm ch}_i(S_t)$ for charger allocation and $l^{\rm tr}_i(S_t)$ for trip assignment. The larger the value of $l^{\rm ch}_i(S_t)$ or $l^{\rm tr}_i(S_t)$, the higher the priority of selecting the EB $i$ for charging or trip assignment. Furthermore, parameter sharing can be applied across the agent networks, with the index $i$ included in the input to distinguish between agents. \par

Next, the logits $\{l^{\rm ch}_i(S_t)\}_{i=1}^M$ for charger allocation and $\{l^{\rm tr}_i(S_t)\}_{i=1}^M$ for trip assignment from all agent networks are fed into the mapping networks for $\omega_t$ and $k_t$, respectively, to generate the charger allocation policy $\mu(\omega_t|S_t)$ and the trip assignment policy $\mu(k_t|S_t)$. Since $\omega_t$ and $k_t$ are independent, and by definition, $o_t = \{\omega_t, k_t\}$, the high-level policy $\mu(o_t|S_t)$ is given by:
\begin{equation}
    \mu(o_t|S_t)=\mu(\omega_t|S_t)\cdot\mu(k_t|S_t).
\end{equation}


The structure of the mapping networks for both $\omega_t$ and $k_t$ consists of three layers: Mask, SoftMax, and Combination/Permutation. 
\begin{itemize}
    \item \textbf{Mask}: Only EBs staying at the terminal station with $B_{i,t}=1$ are eligible for charger allocation or assignment to a new trip. Therefore, the mask layer helps avoid sampling invalid high-level actions. In order to filter out the logits corresponding to the invalid actions, we use a large negative number $\aleph$ (e.g., $\aleph=-1\times10^8$) to replace these logits. The mask sublayer is expressed as 
    \begin{equation}
        \label{masklayer}
        \mathrm{mask}\left( l_{i}^{\mathrm{ch}}(S_t) \right) =\begin{cases}
l_{i}^{\mathrm{ch}}(S_t)&		\mathrm{if}\, B_{i,t}=1\\
	\aleph&		\mathrm{if}\, B_{i,t}=0\\
\end{cases}
    \end{equation}
    \noindent and \eqref{masklayer} also holds for trip assignment by replacing $l_{i}^{\mathrm{ch}}$ with $l_{i}^{\mathrm{tr}}$. 

\item \textbf{SoftMax}: For each EB $i\in\mathcal{M}$, let $p_i^{\rm ch}(S_t)$ represent the probability of assigning EB $i$ to a charger in state $S_t$, i.e.,
\begin{equation}
    \label{p^ch_i}
p_{i}^{\mathrm{ch}}(S_t)=\mathrm{Pr}\left( \omega _{i,t}=1|S_t \right),
\end{equation}
\noindent where $\sum_{i=1}^Mp_{i}^{\mathrm{ch}}(S_t)=1$.

Similarly, let $p_i^{\rm tr}(S_t)$ represent the probability that EB $i$ will be assigned the earliest future trip $\hat{k}_t$ in state $S_t$, i.e.,
\begin{equation}
  \label{p^tr_i}
p_{i}^{\mathrm{tr}}(S_t)=\mathrm{Pr}\left( k_{i,t}=\hat{k}_t|S_t \right),
\end{equation}
\noindent where $\sum_{i=1}^Mp_{i}^{\mathrm{tr}}(S_t)=1$.

The SoftMax layers for $\omega_t$ and $k_t$ take in the masked logits $\{\mathrm{mask}(l^{\rm ch}_i(S_t))\}_{i=1}^M$ and $\{\mathrm{mask}(l^{\rm tr}_i(S_t))\}_{i=1}^M$ from all the agents and convert them into the corresponding probabilities $\{p_{i}^{\mathrm{ch}}(S_t)\}_{i=1}^M$ and $\{p_{i}^{\mathrm{tr}}(S_t)\}_{i=1}^M$, respectively. Specifically, we have
\begin{equation}
    \label{softmax}
    \left\{ p_{i}^{\mathrm{ch}}(S_t) \right\} _{i=1}^{M}=\mathrm{SoftMax}\left( \left\{ {\rm mask}(l _{i}^{\mathrm{ch}}(S_t) )\right\} _{i=1}^{M} \right) ,
\end{equation}
   \noindent and \eqref{softmax} also holds for the trip assignment, by replacing $l_{i}^{\mathrm{ch}}$ and $p_{i}^{\mathrm{ch}}$ with $l_{i}^{\mathrm{tr}}$ and $p_{i}^{\mathrm{tr}}$, respectively.

Due to the Mask layer, both $p_i^{\rm ch}(S_t)$ and $p_i^{\rm tr}(S_t)$ for EBs not at the terminal station are forced to be nearly zero, ensuring that these EBs are not selected.%

\item  \textbf{Combination}:
The SoftMax layer for $\omega _t$ generates the probability of assigning each EB to a charger, ensuring that the probabilities $\{p_{i}^{\mathrm{ch}}(S_t)\}_{i=1}^M$ sum to $1$. Based on these probabilities, we then determine the charger allocation action $\omega_t$. Note that $\omega_t$ corresponds to a combination of $N$ EBs, $C_N=\{ i_1,i_2,\dots,i_n,\ldots ,i_N \}$, where $i_{n}\in \mathcal{M}_t^{\rm lay}$ and each $i_{n}$ is distinct for $n\in\{1,\ldots,N\}$. Accordingly, only the EBs in $C_N$ are allocated chargers, i.e.,   
\begin{equation}
     \omega _{i,t}=\begin{cases}
	1&		\mathrm{if} \,i\in C_N\\
	0&		\mathrm{if} \,i\notin C_N\\
\end{cases}, \forall i \in \mathcal{M}_t^{\rm lay}.
\end{equation}

Therefore, our task is to assign $N$ EBs to chargers from the set of $M_t$ EBs, based on the probabilities $\{p_{i}^{\mathrm{ch}}(S_t)\}_{i=1}^M$. Algorithm \ref{sample_charger} is proposed for this purpose. Consequently, the charger allocation policy can be expressed based on the sequential allocation probability formula as
\begin{equation}
\label{comb}
\mu \left( \omega_t|S_t \right) 
=\sum_{\Pi_{C_N}}\prod_{n =1}^N{\frac{p_{i_{n}}^{\mathrm{ch}}(S_t)}{1-\sum_{\iota =1}^{n -1}{p_{i_{\iota}}^{\mathrm{ch}}(S_t)}}},
\end{equation}
\noindent where $\Pi_{C_N}$ represents the $N!$ permutations of the combination $C_N$. \par





\begin{algorithm}[t]
	\caption{Derive the combination $C_N$}
	\label{sample_charger}
	\begin{algorithmic}[1]  
 \Input 
 \Descb{$\{p_{i}^{\mathrm{ch}}(S_t)\}_{i=1}^M$}{The probabilities of allocating a charger to each EB $i$  }
 \EndInput
 \Output 
 \Descc{$C_N=\{ i_1,i_2,\ldots ,i_N \}$}{The set of $N$ EBs allocated to chargers}
 \EndOutput
 
		\State Initialize $C_N=\{\}$ as an empty set. Initialize the set of remaining EBs $S^{\rm re}=\{1,2,\ldots,M\}$. 
  \For{$n=1$ to $N$} 
  \State Calculate the sum of the probabilities from $S^{\rm re}$ by $p^{\mathrm{sum}}=\sum_{i\in S^{\mathrm{re}}}^{}{p_{i}^{\mathrm{ch}}(S_t)}$.
  \For{$\iota \in S^{\mathrm{re}}$}
  \State Normalize the probability by $p_{\iota}^{\mathrm{ch}}(S_t)\gets p_{\iota}^{\mathrm{ch}}(S_t)/p^{\mathrm{sum}}$
  \EndFor
  \State Sample an EB $i_n$ from $S^{\mathrm{re}}$ based on $\left\{ p_{i}^{\mathrm{ch}}(S_t) \right\} _{i\in S^{\mathrm{re}}}^{}$
  \State Update the combination by $C_N\gets C_{N}\cup \left\{ i_{n} \right\} $
  \State Remove the element $i_n$ from $S^{\mathrm{re}}$
  \EndFor
	\end{algorithmic}  
\end{algorithm}


\begin{algorithm}[t]
	\caption{Derive the permutation $P(\mathcal{M}_t^{\rm lay})$}
	\label{sample_trip}
	\begin{algorithmic}[1]  
  \Input 
 \Descb{$\{p_{i}^{\mathrm{tr}}(S_t)\}_{i=1}^M$}{The probabilities of assigning the earliest future trip $\hat{k}_t$ to each EB $i$}
 \EndInput
 \Output 
 \Descd{$P(\mathcal{M}_t^{\rm lay})=( j_1,j_2,\ldots ,j_{M_t} )$}{The permutation of $M_t$ EBs in the layover period.}
 \EndOutput
 
		\State Initialize $P(\mathcal{M}_t^{\rm lay})=(\,)$ as an empty list. Initialize the set of remaining EBs $S^{\rm re}=\{1,2,\ldots,M\}$. 
  \For{$m=1$ to $M_t$} 
  \State Calculate the sum of the probabilities from $S^{\rm re}$ by $p^{\mathrm{sum}}=\sum_{i\in S^{\mathrm{re}}}^{}{p_{i}^{\mathrm{tr}}(S_t)}$.
  \For{$\iota \in S^{\mathrm{re}}$}
  \State Normalize the probability by $p_{\iota}^{\mathrm{tr}}(S_t)\gets p_{\iota}^{\mathrm{tr}}(S_t)/p^{\mathrm{sum}}$
  \EndFor
  \State Sample an EB $j_m$ from $S^{\mathrm{re}}$ based on $\left\{ p_{i}^{\mathrm{tr}}(S_t) \right\} _{i\in S^{\mathrm{re}}}^{}$
  \State Append $j_m$ at the end of the list $P(\mathcal{M}_t^{\rm lay})$
  \State Remove the element $j_m$ from $S^{\mathrm{re}}$
  \EndFor
	\end{algorithmic}  
\end{algorithm}

\item  \textbf{Permutation}: Based on the probabilities $\{p_{i}^{\mathrm{tr}}(S_t)\}_{i=1}^M$ generated from the SoftMax layer, we should determine the trip assignment action $k_t$. Firstly, as mentioned in Section III.C, $k_t$ can be mapped from the permutation $P(\mathcal{M}_t^{\rm lay})$. Therefore, our task becomes determining the permutation $P(\mathcal{M}_t^{\rm lay})$ based on the probabilities $\{p_{i}^{\mathrm{tr}}(S_t)\}_{i=1}^M$. Algorithm \ref{sample_trip} is proposed for this purpose. Consequently, the trip assignment policy can be expressed based on the sequential allocation probability formula as
\begin{equation}
\label{Permutation}
\mu \left( k_t|S_t \right) 
=\prod_{m =1}^{M_t}{\frac{p_{j_{m}}^{\mathrm{tr}}(S_t)}{1-\sum_{\iota =1}^{m -1}{p_{j_{\iota}}^{\mathrm{tr}}(S_t)}}}.
\end{equation}



Since both \eqref{comb} and \eqref{Permutation} rely on addition and multiplication operations, they maintain differentiability, ensuring that the high-level actor network remains trainable.
\end{itemize}

\subsubsection{State Dimensionality Reduction via Attention Mechanism}

\begin{figure}[t]
\centering
\includegraphics[height=4.0cm,width=0.9\linewidth]{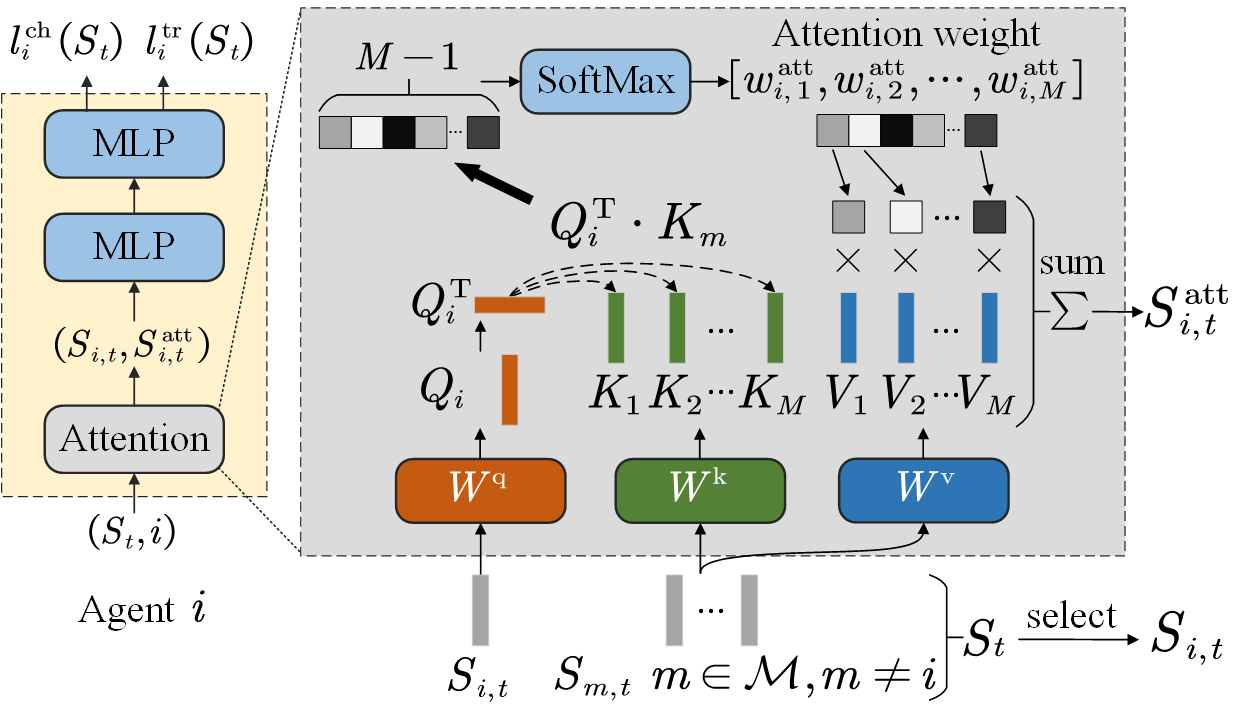}
\caption{The attention layer is utilized to compress the global state $S_t$ to $(S_{i,t},S_{i,t}^{\rm att})$.}
\label{attention} 
\end{figure}

To address the challenge of the high-dimensional global state $S_t$ caused by the large number of EBs, we introduce the attention mechanism to reduce the dimensionality of the input to the high-level actor network.

Specifically, an \textbf{Attention} layer is incorporated into each agent network of the high-level actor.  As shown in Fig. \ref{attention}, the attention layer is placed before the MLP layers, directly processing the input for each agent, i.e., $(S_t,i)$. After passing through the attention layer, the output is $(S_{i,t},S_{i,t}^{\rm att})$, where $S_{i,t}$ is the local state of EB $i$ and $S_{i,t}^{\rm att}$ captures the key features of all other agents' aggregated states that are relevant to EB $i$. Specifically, $S_{i,t}^{\rm att}$ is the weighted sum of other EBs' local states.

The process of obtaining $S_{i,t}^{\rm att}$ involves the three core elements of attention mechanism, i.e., “query”, “key”, and “value” \cite{vaswani2017attention}. As illustrated in Fig. \ref{attention}, the three corresponding different matrices $W^{\rm q}$, $W^{\rm k}$, and $W^{\rm v}$ are used to embed the state information of the agents. Specifically, $W^{\rm q}$ transform $S_{i,t}$ into a query vector $Q_i$, i.e.,
\begin{equation}
    Q_i=W^{\mathrm{q}}\cdot S_{i,t},
\end{equation}
\noindent while the local states of the remaining agents $S_{m,t}$, where $m\in \mathcal{M}\backslash\{i\}$, are transformed by the matrix $W^{\rm k}$ to generate the key vectors $K_m$, i.e.,
\begin{equation}
    K_m=W^{\mathrm{k}}\cdot S_{m,t},\forall m\in \mathcal{M}\backslash\{i\},
\end{equation}
\noindent and by the matrix $W^{\rm v}$ to produce the value vectors $V_m$, i.e., 
\begin{equation}
    V_m=W^{\mathrm{v}}\cdot S_{m,t},\forall m\in \mathcal{M}\backslash\{i\}.
\end{equation}

Next, we compute the dot product for each “query-key” pair, and the resulting $M-1$ dot products are fed into the SoftMax layer to obtain the attention weights $w_{i,m}^{\mathrm{att}}$, i.e.,
\begin{equation}
    \left\{ w_{i,m}^{\mathrm{att}} \right\} _{m\in \mathcal{M}\backslash\{i\}}=\mathrm{SoftMax}\left( \left\{ Q_{i}^{\mathrm{T}}\cdot K_m \right\} _{m\in \mathcal{M}\backslash\{i\}} \right) .
\end{equation}
Finally, these attention weights are used to compute the weighted sum of the value vectors $V_m$, resulting in $S_{i,t}^{\rm att}$ as
\begin{equation}
    S_{i,t}^{\mathrm{att}}=\sum_{m\in \mathcal{M}\backslash\{i\}}{w_{i,m}^{\mathrm{att}}\cdot V_m}.
\end{equation}

\subsubsection{Complexity analysis}

In the basic approach, sampling charger allocation actions requires enumerating the probabilities of all possible combinations of $M$ EBs and $N$ chargers, resulting in a computational complexity of $O(\frac{M!}{N!(M-N)!})$. Meanwhile, the enhanced approach sequentially samples charger allocation actions from $\{p_{i}^{\mathrm{ch}}(S_t)\}_{i=1}^M$ using Algorithm \ref{sample_charger}. The primary computational cost comes from normalizing the remaining probabilities after each sampling step, requiring $\frac{1}{2}\left( 2M-N \right) \left( N-1 \right) $ operations in total, yielding a complexity of $O(M\cdot N)$. A similar pattern applies to trip assignment actions. In the basic approach, the complexity arises from enumerating all permutations of $M_t$ EBs, resulting in $O(M_t!)$. In the enhanced approach, trip assignment actions are sequentially sampled from $\{p_{i}^{\mathrm{tr}}(S_t)\}_{i=1}^M$ using Algorithm \ref{sample_trip}. The normalization after each sampling step requires $\frac{1}{2}\left( M_t+1 \right) \left( M_t-1 \right) $ operations, resulting in a computational complexity of $O({M_t}^2)$. Furthermore, the number of neurons in the output layer of the actor network decreases substantially, from $\frac{M!}{N!(M-N)!}$ to $2$, which alleviates the training difficulty and improves scalability.




The incorporation of the attention layer reduces the number of neurons in the input layer of the actor network from $5\cdot M+2$ to $12$, significantly simplifying the feature extraction process. This reduction not only lowers the computational complexity but also enhances the network's efficiency. Moreover, the attention mechanism improves the algorithm's scalability, enabling the high-level actor to adapt seamlessly to changes in the number of EBs, thereby ensuring robust performance in dynamic fleet scenarios.

\section{Numerical Analysis}
In this section, we conduct experiments to evaluate the effectiveness of the proposed algorithm based on real-world data. All the experiments are performed on a Linux server, using Python 3.8 with Pytorch to implement the DRL-based approaches. 

\subsection{Experimental Setup}
The dataset of time-varying electricity prices is collected from the Midcontinent Independent System Operator (MISO) \cite{MISO2023}. Our experiments focus on two scenarios with different numbers of EBs and chargers to investigate scalability and performance. Scenario 1 sets $M=6$ EBs, which is consistent with the typical settings of closely related works \cite{liu2022robust,yan2024mixed}, where the number of EBs ranges from $4$ to $10$. Scenario 2 scales up to $M=20$ EBs, aligning with the large-scale settings in prior studies \cite{bie2021optimization,wang2024assessing}, but differs by allowing adjustable charging power instead of simple binary charging decisions.

\begin{itemize}
\item 	\textbf{Scenario 1:} We consider $M=6$ EBs and $N=3$ chargers in the terminal station. The buses operate according to the daily schedules based on real-world data \cite{busschedule} from Guelph, Canada. The operating period for each EB follows two normal distributions, i.e., $\mathcal{N}(50,8)$ during rush hours (7:00-9:00 AM and 5:00-7:00 PM), and $\mathcal{N}(40,8)$ during other hours. Each time step is set to $\Delta t=10 \ \mathrm{min}$.
\item 	\textbf{Scenario 2:} We extend the system model to $M=20$ EBs and $N=10$ chargers, maintaining the same operating conditions and time distribution as Scenario 1.
\end{itemize}

We compare the performance of the proposed DAC-MAPPO and DAC-MAPPO-E algorithms with three baseline algorithms in our experiments, including two non-DRL algorithms and one DRL algorithm. 
\begin{enumerate}
    \item \emph{MILP under deterministic setting (MILP-D)}: Our problem under deterministic setting is formulated as a MILP model, where electricity prices and travel times for all EBs are assumed to be known, as described in \cite{manzolli2022electric}. The model is solved using a commercial mixed-integer programming solver, serving as the oracle solution and providing a benchmark for optimal performance, as it operates with perfect information.
    \item \emph{MILP under stochastic setting (MILP-S)}: This method considers the same stochastic setting as our problem, treating electricity prices and travel times as unknown and stochastic variables. To formulate and solve the problem using a MILP solver, travel times are estimated by applying K-means clustering to historical data, using features including EB ID, trip ID, and the trip's departure time, as described in \cite{he2023battery}. For electricity prices, we divide a day into four intervals, i.e., (10:00-15:00), (18:00-21:00), (7:00-10:00, 15:00-18:00, 21:00-23:00), and (0:00-7:00, 23:00-24:00), with a fixed electricity price assigned to each period to align with existing studies, such as \cite{he2023battery}. The price for each interval is estimated based on the average values of historical data from the corresponding period during the week preceding the test day. 
    \item \emph{PPO-MILP}: This approach employs a hybrid hierarchical architecture integrating DRL and MILP, similar to the methods in \cite{yan2024mixed}. At the high level, the DRL algorithm PPO is used to make charger allocation and trip assignment decisions at a fixed interval of every 30 minutes. Based on these high-level decisions, the low-level agent determines the charging power at each time step by solving a MILP. In formulating the MILP, the hourly electricity prices are estimated based on the average values of historical data from the corresponding period during the week preceding the test day. Compared to our fully integrated DRL approach, DRL is only used at the high level to learn the policy over options, without addressing the terminal conditions. Moreover, the low-level intra-option policy, derived from the MILP with estimated electricity prices, is susceptible to prediction errors.\par

\end{enumerate}
MILP-S and PPO-MILP adopt different methods for estimating the electricity prices, in accordance with their respective references \cite{he2023battery} and \cite{yan2024mixed}.\par
The total number of training episodes is set to $20,000$ for Scenario 1 and $30,000$ for Scenario 2. The entire training process for Scenario 1 took approximately 10 hours, while Scenario 2 took around 15 hours. However, once training is complete, the proposed algorithm can make decisions quickly in real time during the deployment phase. The hyper-parameters of the used neural networks are listed in Table \ref{DRLparameters}. Since the high-level actor network in DAC-MAPPO is centralized, we choose a larger high-level actor network size to enable it to learn a more complex policy over options.
\begin{table}[t]
\centering
\caption{Hyper-parameters of various DRL algorithms.}
\begin{tabular}[b]{p{1.2cm}<{\centering}p{1.2cm}<{\centering}p{1.2cm}<{\centering}p{1.2cm}<{\centering}p{0.8cm}<{\centering}p{0.7cm}<{\centering}}
\hline
\textbf{Algorithms}&\textbf{Actor Network Size}&\textbf{Critic Network Size}&\textbf{Beta Network Size}& \textbf{Learning Rate}& \textbf{Batch Size}   \\
\specialrule{0.05em}{0pt}{0pt}
\specialrule{0em}{1pt}{1pt}
\multicolumn{3}{l}{\textbf{DAC-MAPPO}}&&&\\
\specialrule{0em}{1pt}{1pt}
high-level   &128,128&\textbackslash{} &64,64 &3e-4  &128 \\
\specialrule{0em}{1pt}{1pt}
low-level    &64,64&128,128 & \textbackslash{}& 3e-4 & 128  \\		
\specialrule{0em}{1pt}{1pt}
\hline

\specialrule{0em}{1pt}{1pt}
\multicolumn{3}{l}{\textbf{DAC-MAPPO-E}}&&&\\
\specialrule{0em}{1pt}{1pt}
high-level   &64,64&\textbackslash{} &64,64 &3e-4  &128 \\
\specialrule{0em}{1pt}{1pt}
low-level    &64,64&128,128 & \textbackslash{}& 3e-4 & 128  \\		
\specialrule{0em}{1pt}{1pt}

\hline
\specialrule{0em}{1pt}{1pt}
\multicolumn{3}{l}{\textbf{PPO-MILP}}&&&\\
\specialrule{0em}{1pt}{1pt}
high-level   &64,64&128,128 &\textbackslash{} &1e-3  &64 \\
\specialrule{0em}{1pt}{1pt}
\hline
\end{tabular}
\label{DRLparameters}
\end{table}
\subsection{Experimental Results}

\subsubsection{Performance for the test set}
We select data from three different months, i.e., January, May, and September of 2023, for training and evaluation over three runs. For each month, the last week's data is reserved for testing, while the remaining data is used for training. In each run, we execute $100$ complete test episodes and obtain the individual performance of each run by averaging its returns over the test episodes, where the return of one episode is defined as the sum of rewards in each time step, with the reward function given in \eqref{rewardEB}. Table \ref{result} summarizes the individual performances of each run, as well as the average and maximum performances over the three runs for Scenarios 1 and 2. 

\begin{table*}[ht]
\small
\centering
\caption{The individual, average, and maximum performances of all the algorithms across three runs. The performances are derived by averaging the returns over $100$ test episodes, where the return of one episode is defined as the sum of rewards in each time step, with the reward function given in \eqref{rewardEB}.}
\label{result}
\begin{tabular}{|c|l|ccccc|}
\hline
\multirow{2}{*}{\textbf{Scenarios}} & \multirow{2}{*}{\textbf{Algorithms}}    & \multicolumn{5}{c|}{\textbf{Performance}}                                                                                                \\ \cline{3-7} 
                          &                                & \multicolumn{1}{c|}{\textbf{Run 1}}  & \multicolumn{1}{c|}{\textbf{Run 2}}  & \multicolumn{1}{c|}{\textbf{Run 3}}  & \multicolumn{1}{c|}{\textbf{Max}}    & \textbf{Average} \\ \hline
\multirow{5}{*}{1}        & MILP-D      & \multicolumn{1}{c|}{-15.76} & \multicolumn{1}{c|}{-15.47} & \multicolumn{1}{c|}{-15.45} & \multicolumn{1}{c|}{-15.45} & -15.56  \\ \cline{2-7} 
                          & MILP-S & \multicolumn{1}{c|}{-17.83} & \multicolumn{1}{c|}{-17.30} & \multicolumn{1}{c|}{-16.91} & \multicolumn{1}{c|}{-16.91} & -17.35  \\ \cline{2-7} 
                          & PPO-MILP                        & \multicolumn{1}{c|}{-16.90} & \multicolumn{1}{c|}{-16.73} & \multicolumn{1}{c|}{-16.64} & \multicolumn{1}{c|}{-16.64} & -16.76  \\ \cline{2-7} 
                          & DAC-MAPPO              & \multicolumn{1}{c|}{-15.83} & \multicolumn{1}{c|}{-15.59} & \multicolumn{1}{c|}{-15.50} & \multicolumn{1}{c|}{-15.50} & -15.64  \\ \cline{2-7} 
                          & DAC-MAPPO-E           & \multicolumn{1}{c|}{-15.79} & \multicolumn{1}{c|}{-15.56} & \multicolumn{1}{c|}{-15.49} & \multicolumn{1}{c|}{-15.49} & -15.61  \\ \hline
\multirow{5}{*}{2}        & MILP-D      & \multicolumn{1}{c|}{-45.66} & \multicolumn{1}{c|}{-45.49} & \multicolumn{1}{c|}{-45.30} & \multicolumn{1}{c|}{-45.30} & -45.48  \\ \cline{2-7} 
                          & MILP-S & \multicolumn{1}{c|}{-50.35} & \multicolumn{1}{c|}{-49.93} & \multicolumn{1}{c|}{-49.67} & \multicolumn{1}{c|}{-49.67} & -49.98  \\ \cline{2-7} 
                          & PPO-MILP                       & \multicolumn{1}{c|}{-49.88} & \multicolumn{1}{c|}{-49.56} & \multicolumn{1}{c|}{-49.32} & \multicolumn{1}{c|}{-49.32} & -49.59  \\ \cline{2-7} 
                          & DAC-MAPPO              & \multicolumn{1}{c|}{-49.95} & \multicolumn{1}{c|}{-48.90} & \multicolumn{1}{c|}{-49.53} & \multicolumn{1}{c|}{-48.90} & -49.46  \\ \cline{2-7} 
                          & DAC-MAPPO-E           & \multicolumn{1}{c|}{-45.78} & \multicolumn{1}{c|}{-45.53} & \multicolumn{1}{c|}{-45.36} & \multicolumn{1}{c|}{-45.36} & -45.56  \\ \hline
\end{tabular}
\end{table*}


The performance rankings are consistent across both scenarios. MILP-D, representing the theoretical optimal solution, achieved the best average performance. The proposed DAC-MAPPO-E algorithm closely followed, with performance only 0.32\% and 0.18\% lower than MILP-D in Scenarios 1 and 2, respectively. In Scenario 1, DAC-MAPPO exhibited a slightly lower average performance than DAC-MAPPO-E by 0.19\%, whereas in Scenario 2, its performance lagged significantly, with a 8.56\% difference. PPO-MILP showed average performances that were 7.37\% and 8.85\% lower than DAC-MAPPO-E in Scenarios 1 and 2, respectively. Lastly, MILP-S recorded the lowest average performance in both scenarios, trailing DAC-MAPPO-E by 11.15\% in Scenario 1 and 9.70\% in Scenario 2. \par

The fact that both DAC-MAPPO and DAC-MAPPO-E significantly outperform MILP-S demonstrates the effectiveness of DRL approaches in handling uncertainty. The main limitation of MILP-S is the absence of adaptability since it makes decisions based on forecasted travel time/energy consumption and electricity prices. When actual conditions deviate from the predictions, the precomputed solutions become suboptimal. In contrast, as DRL-based approaches, the DAC-MAPPO and DAC-MAPPO-E algorithms are inherently adaptive, continuously updating their decisions based on real-time state information, making them better suited for dynamic environments. \par

Next, while all three DRL-based algorithms adopt a hierarchical structure, PPO-MILP applies DRL only at the high level to handle uncertainties, while its low-level decision-making mirrors MILP-S by ignoring the uncertainty of electricity prices. As a result, its performance remains constrained by the accuracy of forecasted electricity prices. Furthermore, PPO-MILP conducts high-level decision-making at fixed intervals (every $30$ minutes), whereas DAC-MAPPO-E dynamically updates high-level decisions over variable time periods according to the termination function learned through trial and error based on real-world data. This flexible and data-driven approach allows DAC-MAPPO-E to make more efficient decisions, whereas fixed-interval decision-making in PPO-MILP may overlook better alternatives in rapidly changing environments.

Finally, it can be observed that DAC-MAPPO and DAC-MAPPO-E achieve comparable maximum and average performance in Scenario 1, whereas DAC-MAPPO-E significantly outperforms DAC-MAPPO in Scenario 2. This is because the increased number of EBs in Scenario 2 results in higher-dimensional high-level state and action spaces, as well as more complex decision-making requirements. This outcome underscores the importance of the improvements made to the high-level actor network in achieving scalability, as discussed in Section V.B. These enhancements also enable faster convergence during learning and allow the optimal solution to be reached more efficiently, as will be elaborated in Section VI.B.2).

\subsubsection{Convergence Properties}

\begin{figure}[t]
\centering
\subfigure[Scenario 1]{
\label{chargingMAPPO} 
\includegraphics[height=4.0cm,width=0.4\textwidth]{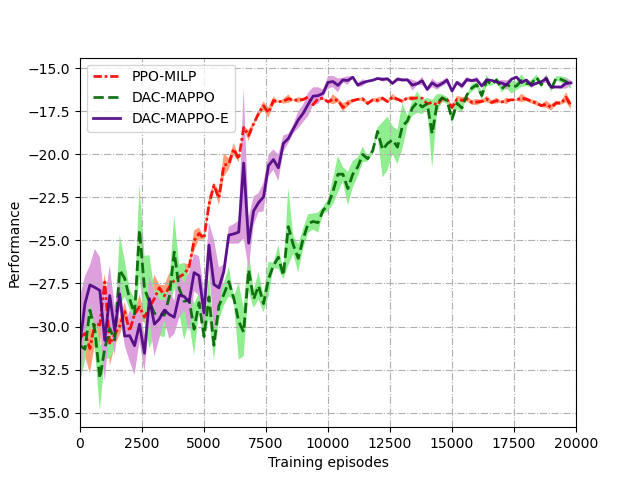}}
\subfigure[Scenario 2]{
\label{chargingCADE} 
\includegraphics[height=4.0cm,width=0.4\textwidth]{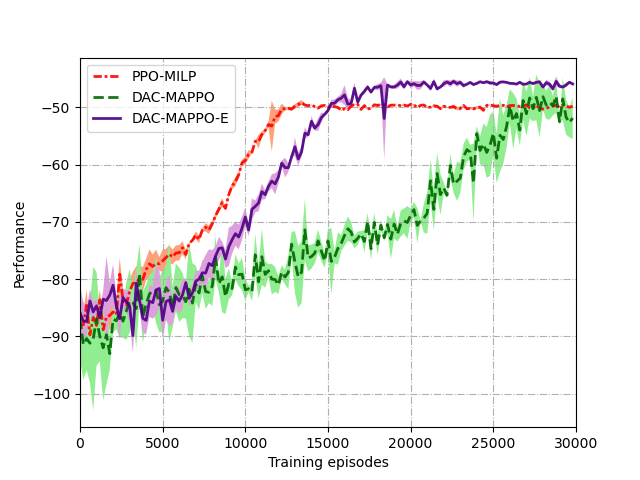}}
\caption{The performance curves of DRL-based algorithms. The shaded areas represent the standard errors across three runs.}
\label{training-result} 
\end{figure}

Fig. \ref{training-result} illustrates the performance curves of three DRL-based algorithms, obtained by periodically evaluating the policies during training. For every $100$ training episodes, $10$ test episodes were conducted, with the X-axis showing the number of training episodes and the Y-axis representing the average performance over $10$ test episodes. The shaded areas indicate the standard errors across the three runs. 

Notably, in Scenario 1, PPO-MILP demonstrates the fastest convergence, stabilizing around $7,500$ episodes. This fast convergence is attributed to its relatively simple architecture, as it only needs to learn the high-level policy using DRL. However, this simplicity comes at the cost of adaptability, as the converged performance of PPO-MILP is worse than the other two algorithms. DAC-MAPPO-E follows PPO-MILP in convergence speed, reaching convergence at approximately $10,000$ episodes, whereas DAC-MAPPO requires considerably more time, converging at about $16,000$ episodes. This indicates that the enhancements in DAC-MAPPO-E improve convergence speed without compromising optimality.  

Meanwhile, in Scenario 2, the convergence patterns reveal notable differences among the algorithms. PPO-MILP and DAC-MAPPO-E converge at approximately $13,000$ and $18,000$ episodes, respectively, while DAC-MAPPO struggles to converge even after $30,000$ episodes. This observation highlights the challenges posed by the increased number of EBs and the resultant expanded state and action spaces to the HDRL algorithms, underscoring the necessity of the enhanced design in DAC-MAPPO-E. Additionally, the shaded area of DAC-MAPPO-E is very small after convergence in both scenarios, indicating our proposed algorithm's stable performance across runs.

\subsubsection{Charging Schedule Results}

\begin{figure*}[t!]
	\centering 
	\subfigure[DAC-MAPPO-E]{
		\includegraphics[width=0.25\linewidth]{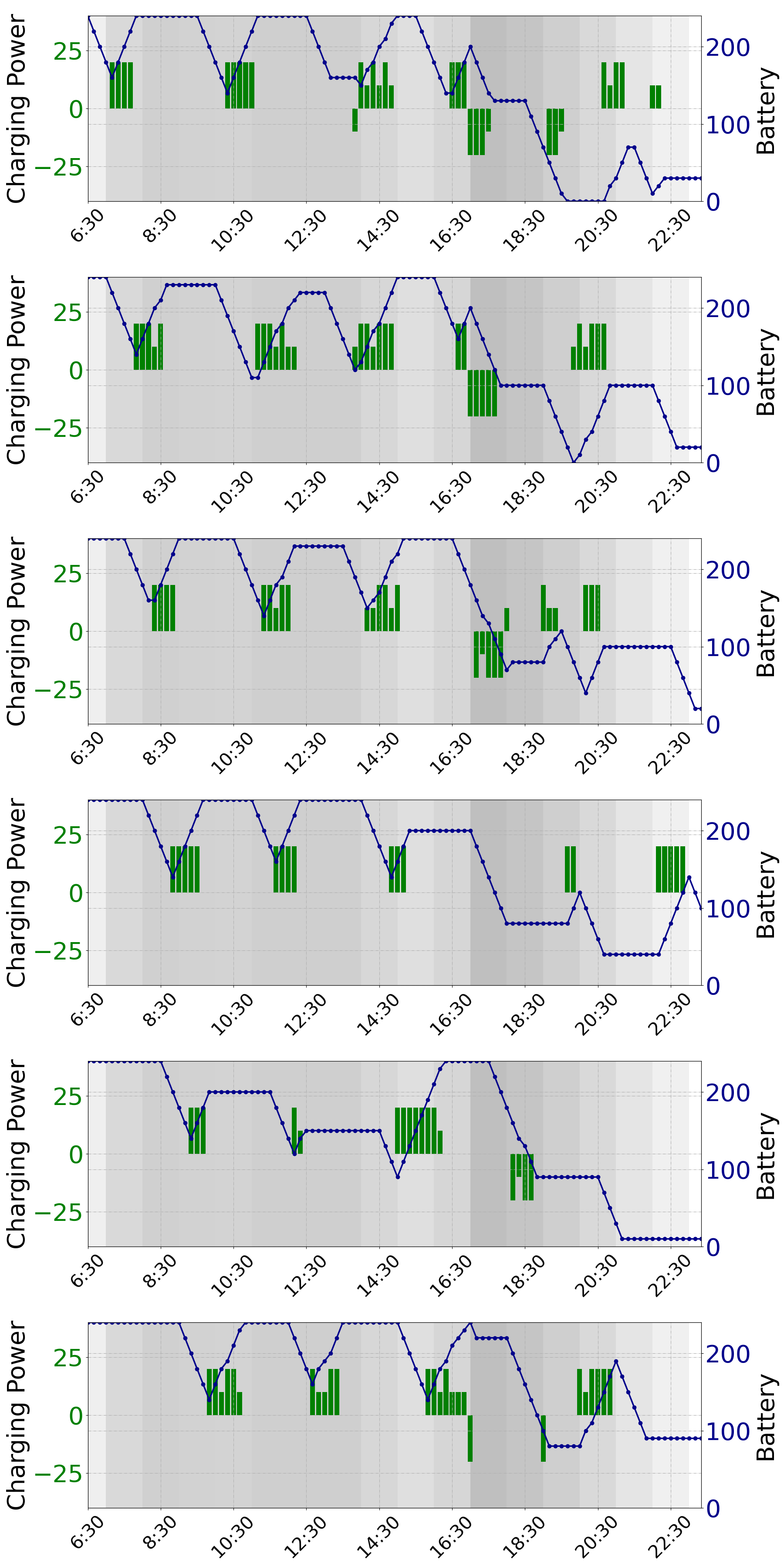}
		\label{trajectoryMAPPO}
	}	
	\subfigure[MILP-S]{
		\includegraphics[width=0.25\linewidth]{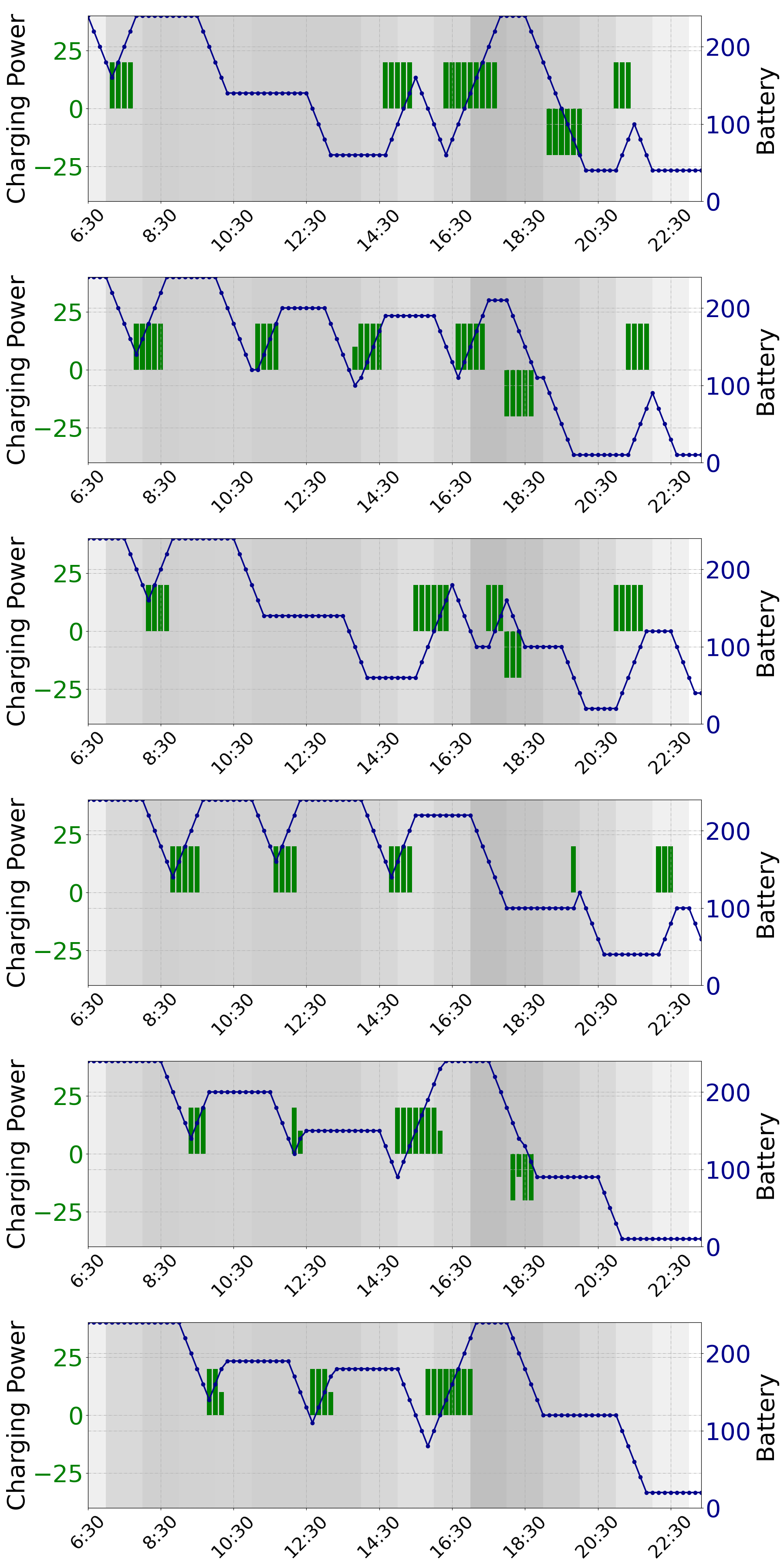}
		\label{trajectoryMILP}
	}
	\subfigure[PPO-MILP]{
		\includegraphics[width=0.25\linewidth]{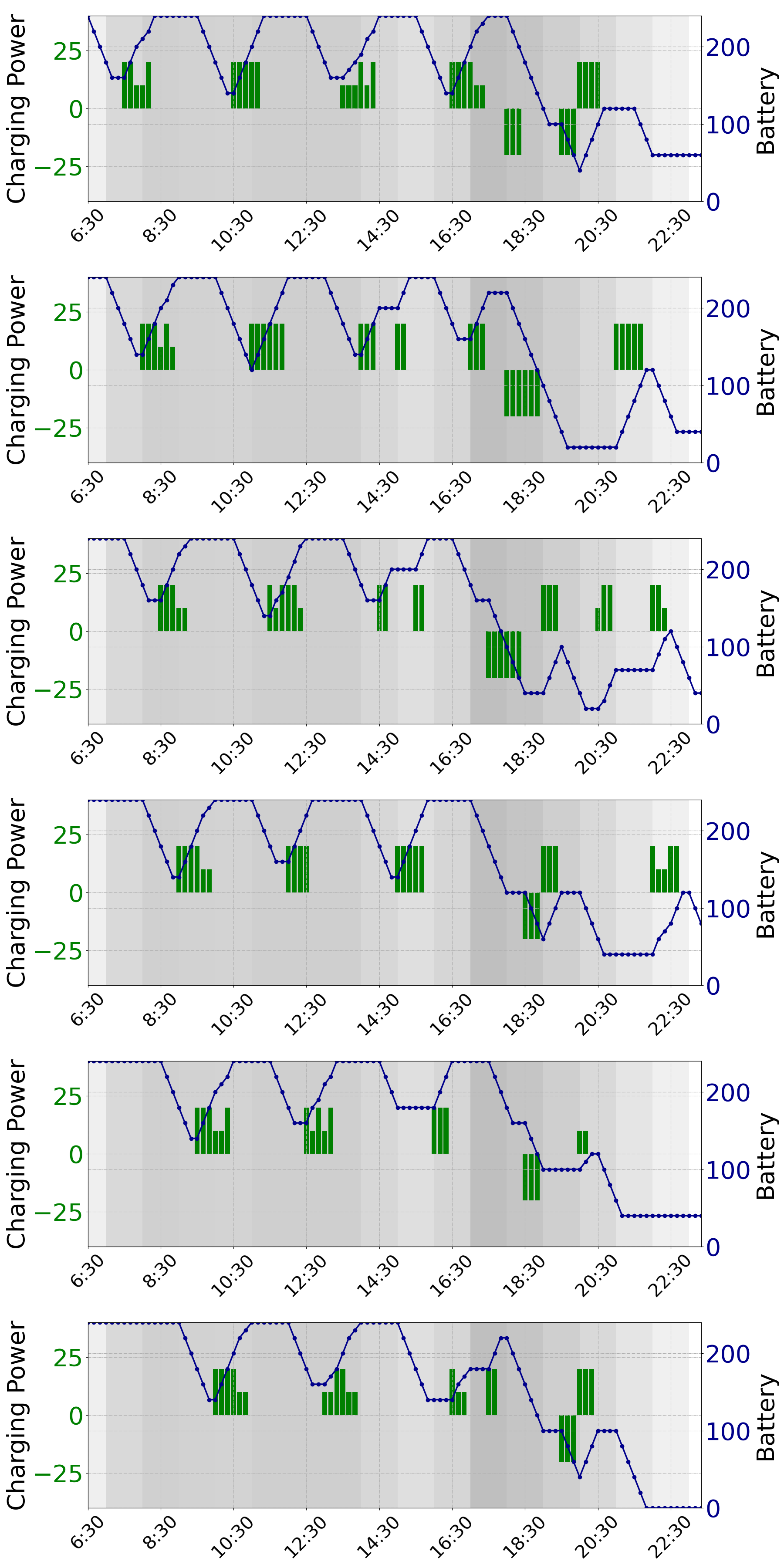}
		\label{trajectoryPPOMILP}
	}\par
	\caption{The detailed charging schedules of DAC-MAPPO-E, MILP-S, and PPO-MILP for Scenario 1.}
    \label{trajectory}
\end{figure*}

To gain insights into the behavior of different charging scheduling policies, we present the charging schedules at each time step for a representative episode from the test set in Scenario 1. Fig. \ref{trajectory} illustrates the results for six EBs under MILP-S, PPO-MILP, and DAC-MAPPO-E, respectively. In these figures, the curves show the battery SoC trajectories of each EB over time, while the bars represent the specific charging power decisions at each time step. Furthermore, the grayscale background highlights electricity price variations, with darker shades indicating higher prices.

Firstly, DAC-MAPPO-E demonstrates a clear ability to optimize charging strategies for maximum profitability. As shown in Fig. \ref{trajectoryMAPPO}, the algorithm strategically increases charging during periods of low electricity prices, such as from 14:00 to 16:00, and capitalizes on price peaks, notably from 17:00 to 18:00, by selling electricity back to the grid. In contrast, MILP-S adopts a less effective strategy that poorly aligns with electricity price fluctuations. For instance, as depicted in Fig. \ref{trajectoryMILP}, EBs A, B, and C continue charging during a high-price period from 17:00 to 18:00, resulting in significantly higher overall costs. This suboptimal performance can be attributed to MILP-S's dependence on forecasted price data, which may deviate considerably from actual values. By comparison, DAC-MAPPO-E leverages the adaptability of DRL, providing it with superior robustness and the capability to handle uncertainties more effectively than MILP-S. The SoC levels of all EBs are low at the end of the day, as they fully recharge overnight at the depot, eliminating the need for energy reservation at the end of the considered time horizon. \par

Next, we compare the charging schedules between PPO-MILP and DAC-MAPPO-E. Similar to MILP-S, PPO-MILP employs MILP at the low level and relies on forecasted electricity price data for optimization. As a result, it also faces the same issue of misaligned charging power decisions with electricity price fluctuations, as discussed above. Moreover, at the high level, PPO-MILP makes charger allocation and trip assignment decisions at fixed time intervals, specifically every 30 minutes, which limits its flexibility. For instance, as shown in Fig. \ref{trajectoryPPOMILP}, EB C returns to the terminal station at 17:10. However, due to the fixed high-level decision interval from 17:00 to 17:30, the agent is unable to allocate a charger to it in time.  Consequently, EB C misses the opportunity to sell more electricity during the peak price period, thereby increasing the overall charging cost to some extent. In contrast, employing the DAC framework, DAC-MAPPO-E can dynamically learn the termination conditions for high-level options, enabling more flexible decision-making for charger allocation and trip assignment. This adaptability helps optimize charging strategies and prevents costly scenarios like those observed with PPO-MILP. 

\section{Conclusion}
In this paper, we have employed HDRL techniques to optimize charging schedules for EB fleets, considering uncertainties in both EB operations and electricity prices. Leveraging the hierarchical architecture of the DAC framework, we have formulated two augmented MDPs to effectively model the EBCSP. Specifically, the high-level charger allocation and trip assignment actions persist over variable time periods, while the low-level charging power actions are selected at each time step. The proposed DAC-MAPPO-E algorithm has successfully solved these augmented MDPs, enabling efficient decision-making across different time scales. The enhancements introduced in DAC-MAPPO-E over the original DAC algorithm span both levels of the hierarchy, leading to improved scalability for managing large-scale fleets. At the low level, integrating the MAPPO algorithm into the DAC framework allows EBs to make local charging power decisions in a decentralized manner, significantly reducing computational complexity and improving convergence speed, particularly with a large number of EBs. At the high level, we have redesigned the actor network structure to substantially decrease the computational complexity of sampling high-level actions and the size of the neural networks. To validate the effectiveness of the proposed algorithm, numerical experiments have been conducted using a real-world dataset. The results have demonstrated the capability of DAC-MAPPO-E in optimizing EB charging schedules efficiently, highlighting its potential for real-world applications. \par

\bibliography{author}
\bibliographystyle{IEEEtran}

\end{document}